\documentclass[entropy,article,accept,moreauthors,pdftex]{Definitions/mdpi}

\firstpage{1} 
\pubvolume{23}
\issuenum{1}
\articlenumber{1640}
\pubyear{2021}
\copyrightyear{2021}
\externaleditor{{Academic Editor: Boris Ryabko} 
}
\datereceived{26 October 2021} 
\dateaccepted{30 November 2021} 
\datepublished{6 December 2021} 
\hreflink{https://doi.org/10.3390/e23121640} 

\usepackage{bbold}
\usepackage{xparse}
\usepackage{physics}
\usepackage{nicefrac, dsfont, morefloats}
\usepackage[labelformat=simple]{subcaption}

\DeclareCaptionLabelFormat{subcaptionlabel}{\normalfont(\textbf{#2}\normalfont)}
\Title{Probabilistic Autoencoder Using Fisher Information}

\TitleCitation{Probabilistic Autoencoder Using Fisher Information}

\Author{Johannes Zacherl\orcidA 
 $^{1, 2}$*, Philipp Frank\orcidB $^{1,2}$ and Torsten A. En{\ss}lin\orcidC $^{1,2}$}

\AuthorNames{Johannes Zacherl, Philipp Frank and Torsten A. En{\ss}lin}

\AuthorCitation{Zacherl, J.; Frank, P.; En{\ss}lin, T.A.}

\address{
$^{1}$ \quad Max-Planck Institut f{\"u}r Astrophysik, Karl-Schwarzschild-Stra{\ss}e 1, 85748 Garching, Germany; {philipp@mpa-garching.mpg.de (P.F.);} 
 ensslin@mpa-garching.mpg.de (T.A.E.)\\
$^{2}$ \quad {Faculty of Physics, Ludwig}
-Maximilians-Universit{\"a}t M{\"u}nchen, Geschwister-Scholl-Platz 1, 80539 M{\"u}nchen, Germany}

\corres{Correspondence: zacherl@mpa-garching.mpg.de}

\abstract{Neural networks play a growing role in many scientific disciplines, including physics. Variational autoencoders (VAEs) are neural networks that are able to represent the essential information of a high dimensional data set in a low dimensional latent space, which have a probabilistic interpretation. In particular, the so-called encoder network, the first part of the VAE, which maps its input onto a position in latent space, additionally provides uncertainty information in terms of variance around this position. In this work, an extension to the autoencoder architecture is introduced, the FisherNet. In this architecture, the latent space uncertainty is not generated using an additional information channel in the encoder but derived from the decoder by means of the Fisher information metric. This architecture has advantages from a theoretical point of view as it provides a direct uncertainty quantification derived from the model and also accounts for uncertainty cross-correlations. We can show experimentally that the FisherNet produces more accurate data reconstructions than a comparable VAE and its learning performance also apparently scales better with the number of latent space dimensions.}

\keyword{machine learning; Fisher information metric; variational methods; variational autoencoder; deep generative network}

\RequirePackage[normalem]{ulem} 
\RequirePackage{color}\definecolor{RED}{rgb}{1,0,0}\definecolor{BLUE}{rgb}{0,0,1} 

\begin{document}

\section{Introduction}

Machine learning has become a key method for data analysis~\cite{Carleo.2019}.
Many machine learning methods can be classified as supervised or unsupervised learning.
In supervised learning, the machine learning model is trained to output a specific feature of the data, for example, to classify the data into some predetermined categories.
To train a supervised model, we rely on training data, which are labeled with the features we want to learn.
The labels need to be attached to the data manually, and therefore, the number of labeled data sets is limited.
In contrast, in unsupervised learning, the machine learning model is trained to learn patterns from the data directly,  without any predetermined features.
Thus for many problems without labeled data available, we rely on unsupervised learning techniques to learn a structure underlying our data.
\par
One important unsupervised learning strategy is generative modeling 
\cite{Lamb.2021}.
Generative models learn an underlying probability distribution of the data by simulating the data generating process that is seeded with an input drawn from a simple, unstructured probability distribution.
In this way, the generative model learns to represent and sample from the complex, underlying distribution.
Generative models allow us to encode and use prior expert knowledge of a given problem~\cite{Knollmuller.2018}.
A number of neural networks are generative models, and we discuss in the following autoencoders (AE)~\cite{Rumelhart.1986}, variational autoencoders (VAE)~\cite{Kingma.2013, Rezende.2014}, and generative adversarial networks (GANs)~\cite{Goodfellow.2014}.
\par
All mentioned models share the concept of a latent space, which often is a lower-dimensional space than the data space, the space on which the data vectors are defined. 
A simple prior probability distribution is postulated for the latent variables. 
The data distribution is reproduced by means of a transformation function from the latent space to the data space. 
This transformation is usually represented in terms of a trainable neural network and thus, after training, provides a generative model.
\par
In an AE, the latent space is the central layer of the network. 
AEs take vectors in the input data space, translate these into the latent space, and then back into the output space, which is isomorphic to the initial data space. Thereby, the latent space is usually of  a lower dimension than the input data space to enforce the network to encode only essential information. 
AEs become optimized to reproduce the input data in the output layer and work without any stochasticity.
\par
The VAE is a deep generative model that expands the idea of an AE with probabilistic methods using latent space inference techniques. 
In particular, encoding a data point in latent space can be regarded as an inference problem given the data generating process described by the decoder. 
The encoder's goal is to infer that latent space point that---if passed through the decoder---would regenerate the original data point. 
In contrast to the AE, a VAE solves this inference problem probabilistically, i.e., via a variational approximation of the latent space posterior distribution with a Gaussian distribution. 
This latent space distribution is parameterized via a mean and a (assumed to be) diagonal covariance, which both are  outputs of the encoder, given a specific input data point.
The encoder and decoder networks are trained jointly.
\par
This training is undertaken by learning the generative function the decoder represents in the form of a transformation from the latent space to the data space and additionally using Bayesian inference to find the latent space representation of the data.
To do this, the simple Gaussian prior distribution for the latent space is updated using Bayes' Theorem with the data in order to find the posterior latent space distribution.
This calculation is often intractable, so one needs to rely on approximations to access the posterior.
\par
VAEs use variational inference~(VI) for this approximation.
In VI, the posterior is approximated by choosing a parametrized probability distribution and fitting it to the true posterior via the variational parameters.
VI-based methods are relatively efficient and provide good approximations for cases where the true posterior distribution can be approximated well by the parametrized distribution.
Since the parametrized distribution is usually  chosen ad hoc, it is not necessarily known whether it is actually capable of providing a good approximation to the true posterior~\cite{Frank.2021}.
For a detailed discussion of VI techniques in generative models, we refer the reader to~\cite{Wainwright.2007}.
\par
To perform variational inference via the encoder, the VAE usually relies on the so-called mean-field approach, which assumes statistical independence between the different dimensions of the latent space.
Typically the parametrized distribution is chosen as a multivariate Gaussian with a diagonal covariance matrix.
Therefore, the variational parameters the encoder network has to provide are the mean and the diagonal of the covariance.
The decoder then reconstructs the data from a sample from the approximate latent space distribution.
\par
The VAE typically covers the data distribution fully but is lacking in the quality of generated samples.
Therefore, other models such as GANs~\cite{Goodfellow.2014} have become more popular for data generation in recent years.
GANs train a generative neural network by producing data that is then  judged by a discriminator network on whether it is a real or generated sample.
This way, the GAN becomes trained to generate new samples of the given data set, indistinguishable from the real data.
While GANs can produce very high-quality samples, they are prone to mode collapse  and hence fail to cover the full data distribution~\cite{Grover.2018, Arora.2017}.
\par
Due to these limitations of GANs, the VAE remains an interesting model, and a number of approaches have been proposed to improve on its weaknesses.
One example is the $\beta$-VAE~\cite{Higgins.2017}.
Here, a prefactor $\beta$ is introduced in the loss function of the VAE in order to balance the regularizing terms.
A drawback to this method is that it requires expensive hyperparameter tuning for the introduced $\beta$ factor.
\par
Other attempts of expanding the VAE framework in different directions include an expansion to dynamical models~\cite{Gregor.2015} and the use of auxiliary latent variables~\cite{Salimans.2015, Ranganath.2016, Maaloe.2016}.
The idea behind inference with auxiliary latent variables is to define and infer a joint distribution over the latent and the auxiliary latent variables, which implies a potentially potent marginal distribution for the latent variables. 
This provides access to a more flexible class of inference distributions.
\par
A different approach towards more flexible VI is the normalizing flow~\cite{Rezende.2015}, where the posterior distribution is built via an iterative procedure.
Normalizing flows describe the data distribution in terms of a simple (usually Gaussian) latent space distribution and apply a series of invertible parametrized transformations to it, which results in an invertible generative network. 
The requirement of invertibility typically implies that the approach does not scale well to high dimensional latent spaces and many different transformations and flows have been proposed to overcome this issue~\cite{Kingma.2016, Germain.2015, vanOord.2016, Dinh.2017}.
  \par
 In order to make the inference process more robust, attempts have been made using different measures than the KL-divergence to perform inference.
To this end, the Wasserstein autoencoder 
\cite{Tolstikhin.2018} 
used the Wasserstein metric, and the Fisher autoencoder 
\cite{Elkhalil.2020} 
relied on the Fisher divergence 
\cite{Ding.2019}.
Hybrid models between VAE and GAN have also been proposed to try and obtain the best from both models~\cite{Dumoulin.2017, Grover.2018, Rosca.2018}.
\par
In this work, we expand the VI approach of a VAE to make it more flexible in a different way to the approaches discussed above.
We use a different inference approach by formulating an approximate joint posterior distribution for the latent space variables, the model parameters of the generative model, and a noise parameter and minimizing the Kullback--Leibler (KL)~\cite{Kullback.1951} divergence to the true posterior.
Additionally, we adapt the inference approach of metric Gaussian variational inference (MGVI)~\cite{Knollmuller.2019}, where the approximate distribution for the latent variables is a multivariate Gaussian, which uses a metric based on the Fisher information metric as the covariance.
Due to its use of the Fisher metric, we call the proposed VAE variant FisherNet.
This approach allows us to account for correlations in latent space in an efficient manner and reduces the number of variational parameters by using the Fisher information metric calculated from the generative process. 
This allows us to improve on some aspects of the standard VAE without requiring additional hyper parameter tuning.
During the derivation of the FisherNet, we make a number of simplifying assumptions and approximations.
At the end of the paper, we provide a list of the assumptions made. This list is informative of the possible limitations of the derived method and may provide hints for future improvements.

\section{Materials and Methods}
In this section, we first derive the Bayesian interpretation of a VAE and afterward introduce the modifications we made to arrive at the FisherNet.
The derivation of the Bayesian VAE interpretation is  analogous to~\cite{Milosevic.2021}.

\subsection{Data Generating Process} \label{sec:data_gerating_process}
Generative models are mappings of parameters that have a simple probability distribution that is thereby mapped to a more complex one. 
Essentially, they are models for a process that takes a random variable and generates a different, more complex random variable.
Generative models are useful since we can use them for mapping  spaces where the priors needed for Bayesian inference techniques are simple, as well as spaces where the priors are more complex.
This mapping provides a convenient way to encode prior knowledge we have about the system into the generative model~\cite{Knollmuller.2018}.
\par
In order to design our generative model, we will start by introducing a noisy data model  $D = \tilde{D} + \tilde{N}$ and a generating function $f_{\theta}(z_i)$, which maps from an abstract latent $\mathcal{Z}$ space to the data space $\mathcal{D}$.
Here, $D = (d_i)_{i=1}^{p}$ is the collection of data vectors $d_i = D_{\cdot i}$, where $i \in \qty{1,...,p}$ and $p$ are the number of data vectors. 
Each data vector has a generated vector $\tilde{d}_i  = f_\theta(z_i)$ and a noise vector $\tilde{n}_i$ associated to it, which is analogously  combined into the sets $\tilde{D} = (\tilde{d}_i)_{i=1}^{p}$ and $\tilde{N} = (\tilde{n}_i)_{i=1}^{p}$.
The generative process is a complex but analytical, nonlinear function parameterized by a set of parameters $\theta$, which are the same for all $i$.
Specifically,

\begin{align}
f\! : \ &\mathcal{Z} \rightarrow \mathcal{D} \nonumber\\\
&z_i \mapsto f_{\theta}(z_i)=\tilde{d_i}\label{data_generating_function}
\end{align}
where $f_{\theta}(z_i)$ generates a k-dimensional data vector from a lower-dimensional latent space vector $z_i = Z_{\cdot i}$.
The latent space vectors are combined in the set $Z=(z_i)_{i=1}^{p}$.
The data are given by the result of the generating function plus noise

\begin{equation}
d_i = f_{\theta}(z_i) + n_i,\label{data_equation}
\end{equation}
where the latent space vectors $z_i$ and the noise $n_i$ are independent.
The goal of our approach will be to find the parameters for this process as well as the target distribution of the data for a given data set.
To this end, we will work within the framework of Bayesian inference.
This means we set priors for the desired parameters and then look for the posterior distribution of those parameters, given the data set $D$.

\subsection{Priors and Likelihood} \label{prior_chapter}
The prior distributions for the different variables and model parameters are the starting point for finding a posterior via Bayesian inference.
For the latent space, we choose a standard Gaussian distribution

\begin{equation}
P(Z)= \prod^p_{i=1} P(z_i) = \prod^p_{i=1} \dfrac{1}{\sqrt{|2\pi \mathds{1}|}} \exp(-\frac{1}{2}z_i^\dagger \mathds{1} z_i) =  \prod^p_{i=1} \mathcal{G}(z_i,\mathds{1}) ,
\label{latent space prior}
\end{equation}
where $\dagger$ indicates the adjoint of a vector, $\mathds{1}$ is the unit matrix, and $|...|$ is a determinant.
We can choose this prior without loss of generality since for every set of latent space vectors we can find a transformation into a coordinate system where they are distributed according to a standard Gaussian distribution.
This transformation is also known as random variable generation with inverse transform sampling~\cite{Devroye.1986} and is the basis for the reparametrization trick~\cite{Kingma.2013, Rezende.2014, Titsias.2014}.
It allows us to solely use the standard Gaussian distribution for our calculations, which significantly simplifies them.
\par
At this point, we have to set one more prior as well as to decide on a model for the noise distribution.
The prior we have to set is the one for the model parameters $\theta$ that parametrizes the generating function $f_\theta(z)$.
For those parameters, we select a uniform distribution.
For the noise, we use a Gaussian distribution

\begin{equation}
P\left(\tilde{N}\right)=\prod^p_{i=1} \mathcal{G}(n_i,N) ,
\end{equation}
where we assume directional independence, { i.e.,} $N=\sigma_n^2 \,\mathds{1}$.
From Equation~\eqref{data_equation}, we can see that the level of noise tells us how accurately we can reconstruct the data with the generative process.
Since the appropriate noise level $\sigma_n^2$ is not necessarily known beforehand, we make it a learnable parameter of the model.
To implement this, we can introduce a single standard Gaussian parameter $\xi_N$ and use a log-normal mapping $t(\xi_N)$ to map it onto the noise covariance,

\begin{equation}
N = t(\xi_N) = \mathds{1}\exp(\xi_N).
\end{equation}
{This}
 noise model is a simplification, as the noise distribution does not need to be the same everywhere in data space, nor isotropic, nor Gaussian.
\par
Now that we have introduced all the priors, the last remaining element we need is the likelihood.
We start with a general likelihood $P(d_i,n_i|z_i,\theta,\xi_N)$, which describes how likely the generating process' output is, given all relevant parameters.
To find the likelihood of the data, we marginalize this distribution over the noise:

\begin{align}
P(d_i|z_i,\theta, \xi_N) &= \int dn_i P(d_i, n_i | z_i, \theta, \xi_N) \\\
&= \int dn_i P(d_i|z_i,\theta,n_i)P(n_i|\xi_N)\\\
&=\int dn_i \delta(d_i-f_{\theta}(z_i)-n_i)\mathcal{G}(n_i,t(\xi_N))\\\
&=\mathcal{G}(d_i-f_{\theta}(z_i),t(\xi_N)) .
\label{likelihood}
\end{align}
With the priors set and the likelihood calculated, we can write down an expression for the posterior distribution, using Bayes' theorem and assuming a priori independence between $Z$, $\theta$, and $\xi_N$:

\begin{equation}
P(Z, \theta,\xi_N|D) = P(D|Z,\theta,\xi_N)\dfrac{P(\theta)P(\xi_N)P(Z)}{P(D)}.
\label{true_posterior}
\end{equation}
This posterior is the distribution we are interested in. 
The datum $D$ is the known parameter for our problem. 
We want to find the parameters $Z$, $\theta$, and $\xi_N$ that specify the generating process.

\subsection{Approximating the Posterior}
Since it is generally not possible to calculate the distribution in Equation~\eqref{true_posterior} analytically, we will approximate it with a second distribution $Q_{\phi}(Z,\theta,\xi_N|D)$ with variational parameters $\phi$.
For the approximation, we also assume a posteriori independence between the parameters $\theta$ and $\xi_N$, in addition to their prior independence. The approximate latent space distribution depends on both $\theta$ and $\xi_N$ via the Fisher metric, which we introduce in the next section. This allows us to write the approximate posterior as
\begin{equation}
Q_{\phi}(Z,\theta,\xi_N|D)=Q_{\phi}(Z|D, \theta, \xi_N)Q_{\phi}(\theta|D)Q_{\phi}(\xi_N|D).
\end{equation}
Assuming independence allows us to choose different strategies for the approximations of the posteriors of the various parameters.
For the model parameters and the noise, we choose a Maximum a posteriori~(MAP) solution.
This allows us to write those estimate distributions as delta functions $Q_{\phi}(\theta|D)=\delta(\theta-\hat{\theta})$ and $Q_{\phi}(\xi_N|D)=\delta(\xi_N-\hat{\xi}_N)$, where $\hat{\theta}$ and $\hat{\xi}_N$ are the respective MAP estimates.
\par
For the latent space variable $z_i$, we want to find a more expressive distribution than a delta distribution.
Therefore, we will perform variational inference for $Q_{\phi}(Z|D, \theta, \xi_N)$.
We model this distribution with a Gaussian 

\begin{equation}
Q_{\phi}(Z|D, \theta, \xi_N)=\prod_i\mathcal{G}(z_i-\mu_i,M^{-1}(\mu_i)),
\end{equation}
where the dependence on $\theta$ and $\xi_N$ are via $M^{-1}(\mu_i)$.
We only use the mean $\mu$ as a variational parameter, which is inferred from the data via the encoder $\mu_i = g_\phi(d_i)$.
As the covariance matrix, we use $M^{-1}(\mu_i)$, a metric based on the Fisher Information metric. 
Putting the three distributions together, the approximate distribution we use for the inference is

\begin{equation}
\label{approximate_posterior}
Q_{\phi}(Z,\theta,\xi_N|D) = \prod_i\mathcal{G}(z_i-\mu_i,M^{-1}(\mu_i)) \delta(\theta-\hat{\theta}) \delta(\xi_N-\hat{\xi_N}).
\end{equation}

\subsection{A Metric as Covariance} \label{sec:fisher_cov}
A VAE approximates the local distribution of the latent space variables for a single data point using a Gaussian distribution with variational paramaters, namely the mean $\mu_i$ and a diagonal covariance.
We change this approach to the inference of the latent space distribution by dropping the covariance as a variational parameter and using $M^{-1}$, a metric based on the Fisher metric that is calculated directly from the generative model instead.
The concept of using this metric to approximate the uncertainty stems from the MGVI algorithm~\cite{Knollmuller.2019}.
As it was shown and becomes clear below, this metric is a strictly positive definite matrix.
Using it as a proxy for the posterior covariance brings about two theoretical advantages.
First, it is a full matrix, i.e., in general non-diagonal, 
 which allows us to consider correlations in latent space and not impose local independence on the latent space variables.
Second, we can calculate it directly from the generative process modeled by the decoder of the FisherNet.
Therefore, we reduce the number of variational parameters and instead utilize the information from the generative part of the model.
For a thorough discussion of the merits of this approximation, we refer the reader to~\cite{Knollmuller.2019, Frank.2021}.
\par
Our approximation metric consists of two additive components $M=I_d + M_z$, which we can calculate.
The first, $I_d$, is the Fisher metric of the likelihood

\begin{equation}
I_d(\tilde{d_i}) = \left\langle \dfrac{\partial \mathcal{H}(d_i|\tilde{d_i})}{\partial \tilde{d_i}}\dfrac{\partial \mathcal{H}(d_i|\tilde{d_i})}{\partial \tilde{d_i}^{\dagger}}\right\rangle_{P(d_i|\tilde{d_i})} .
\end{equation}
Expectation values are written as $\left\langle ... \right\rangle_{P(...)}$.
This form ensures that $I_d(\tilde{d}_i)$ is positive definite.
When inverted, this term corresponds to the Cram\'{e}r--Rao bound~\cite{Cramer.1946,Rao.1992}, which is a lower bound of the uncertainty of an estimator in frequentist statistics and estimation theory:

\begin{equation}
I_d(\tilde{d_i})^{-1} \leq \left\langle\qty(\tilde{d_i} - \left\langle \tilde{d_i}\right\rangle)\qty(\tilde{d_i} - \left\langle \tilde{d_i}\right\rangle)^\dagger\right\rangle_{P(d_i|\tilde{d_i})},
\end{equation}
where the inequality indicates that subtracting the left from the right side of the equation yields a positive semi-definite matrix. 
\par  
To rewrite the Fisher metric in the coordinates of the generative model, we write it as the push forward of the metric in the original parametrization via the generative transformation $\tilde{d_i} = f_\theta(z)$.

\begin{align}
\eval{I_d(z_i)}_{z_i=\mu_i} &= \eval{\left\langle \dfrac{\partial \mathcal{H}(d_i|z_i)}{\partial z_i}\dfrac{\partial \mathcal{H}(d_i|z_i)}{\partial z_i^{\dagger}}\right\rangle_{P(d_i|z_i)}}_{z_i=\mu_i} \\\
&=\left( \dfrac{\partial f(\mu_i)}{\partial \mu_i}\right)^{\dagger}  \left\langle \dfrac{\partial \mathcal{H}(d_i|\tilde{d_i})}{\partial \tilde{d_i}}\dfrac{\partial \mathcal{H}(d_i|\tilde{d_i})}{\partial \tilde{d_i}^{\dagger}}\right\rangle_{P(d_i|\tilde{d_i})} \left( \dfrac{\partial f(\mu_i)}{\partial \mu_i}\right) \\\
&= J(\mu_i)^{\dagger}I_d(f(\mu_i))J(\mu_i) ,
\end{align}
where $J(\mu_i)$ is the Jacobian of the generating function.
Furthermore,  $I_d(z_i)$ is positive definite as a consequence of  $I_d(\tilde{d}_i)$ being positive definite.
\par
The likelihood we found in Equation~\eqref{likelihood} is a Gaussian distribution.
Therefore, the Fisher metric of the likelihood is the inverse covariance of the likelihood, which we found to be the noise covariance $I_d(f(\mu_i))=N^{-1}=t\qty(\xi_N)^{-1}$.
\par
The second component $M_z$ is the Hessian of the prior distribution, which is the identity matrix for the standard Gaussian prior, we defined in Equation~\eqref{latent space prior}.

\begin{equation}
where M_{z} =\frac{\partial \mathcal{H}(z_i)}{\partial z_i  \partial z_i^\dagger} =\mathds{1} .
\end{equation}
$M_z$ is strictly positive definite.
In the case we consider here, the inverse Hessian is equivalent to the uncertainty of the prior distribution.
\par
Combining the two components, we arrive at the metric

\begin{equation}
M_i = M(\mu_i) = J(\mu_i)^{\dagger} t(\xi_N)^{-1}J(\mu_i) + \mathds{1},
\label{eq:Gaussian_fishermetric}
\end{equation}
which is strictly positive definite and therefore invertible.
The inverse of this metric is what we use as the approximation for the latent space posterior distribution's covariance.
It is a lower bound to the actual posterior uncertainty.

\subsection{Kullback--Leibler Divergence}

Since we want our approximate posterior to be as close as possible to the actual posterior, we fit it via variational inference.
Seeing that the true posterior is not available to us, we need to calculate the inference KL between the two distributions $P(Z, \theta,\xi_N|D)$ and $Q_{\phi}(Z,\theta,\xi_N|D)$.
Performing variational inference, the training goal of our model is to minimize this KL divergence.
This means we optimize the variational parameters in order to obtain the approximate distribution as close as possible to the true posterior.
\par
A straightforward calculation of this KL divergence can be found in \citep{Milosevic.2021} and provides:

\begin{align}
\mathcal{D}_\mathrm{KL} \left[ Q\Vert P\right] &= \frac{1}{2}\sum_{i=1}^p\left[ -\ln(|M^{-1}_i|) + \int \dd z_i \mathcal{G}\qty(z_i -\mu_i,M^{-1}_i)\mathrm{tr}\qty(z_iz_i^{\dagger}) \right. \nonumber\\  
 &+\int \dd z_i \mathcal{G}(z_i-\mu_i, M^{-1}_i) \mathrm{tr}\left( t(\hat{\xi}_N)^{-1}(d_i-f_{\hat{\theta}}(z_i))(d_i-f_{\hat{\theta}}(z_i))^T\right) \nonumber\\
 &+\left. \mathrm{tr}\qty( \ln(t(\hat{\xi}_N))) + \dfrac{1}{p}\hat{\xi}_N^2\right] + H_0 .
\end{align}
All constant terms are absorbed into $H_0$.
While it is possible to calculate the first of the remaining integrals in terms of the moments of the posterior distribution $\mu_i$ and $M_i^{-1}$, the same is not true for the second integral.
Since we want to treat the terms equally, we approximate those integrals via

\begin{equation}
\int \dd z_i \mathcal{G}\qty(z_i -\mu_i,M^{-1}_i)\mathrm{tr}\qty(z_iz_i^{\dagger}) \approx \dfrac{1}{S} \sum_{m=1}^S \mathrm{tr}\qty(z_{i}^*z_{i}^{*\dagger}) ,
\end{equation}
and

\begin{align}
&\int \dd z_i \mathcal{G}(z_i-\mu_i, M^{-1}_i) \mathrm{tr}\left( t(\hat{\xi}_N)^{-1}(d_i-f_{\hat{\theta}}(z_i))(d_i-f_{\hat{\theta}}(z_i))^T\right) \nonumber\\
&\approx \dfrac{1}{S} \sum_{m=1}^S \mathrm{tr}\left( t(\hat{\xi}_N)^{-1}(d_i-f_{\hat{\theta}}(z_{i}^*))(d_i-f_{\hat{\theta}}(z_{i}^*))^T\right) ,
\end{align}
where $z_{i}^*$ denotes a sample drawn from the approximative distribution $G(z_i-\mu_i, M^{-1}_i)$ and S is the number of samples used to approximate the result.
We sample once for each input vector, so $S=1$.
This lets us simplify the result of the KL to

\begin{align}
\mathcal{D}_\mathrm{KL} \left[ Q\Vert P\right] &\approx \frac{1}{2}\sum_{i=1}^p\left[ -\ln(|M^{-1}_i|)+\mathrm{tr}\left( z_{i}^*z_{i}^{*T}\right)\right.  \nonumber \\\
&+ \mathrm{tr}\left( t(\hat{\xi}_N)^{-1}(d_i-f_{\hat{\theta}}(z_{i}^*))(d_i-f_{\hat{\theta}}(z_{i}^*))^T\right) \nonumber \\\ 
&+\left. \mathrm{tr}\left( \ln(t(\hat{\xi}_N))\right) +\dfrac{1}{p}\hat{\xi}_N^2 \right] + H_0 \label{Full_KL_divergence}.  
\end{align}

Assuming $M_i$ to be sufficiently constant around the mean allows us to put the respective term into $H_0$. 
This yields

\begin{align}
\mathcal{D}_\mathrm{KL} \left[ Q\Vert P\right] &\approx \frac{1}{2}\sum_{i=1}^p\left[ \tr\left( z_{i}^*z_{i}^{*T}\right)+\dfrac{1}{p}\hat{\xi}_N^2 +  \tr\left( \ln(t(\hat{\xi}_N))\right)\right. \nonumber \\\
&+\left. \tr\left( t(\hat{\xi}_N)^{-1}(d_i-f_{\hat{\theta}}(z_{i}^*))(d_i-f_{\hat{\theta}}(z_{i}^*))^{T}\right)\right] + H_0. \label{eq:loss_function}
\end{align}
This result is the loss function that we minimize in the training process of the FisherNet.
(The loss function of the VAE, the so-called evidence lower bound~(ELBO)~\cite{Kingma.2013}, can be derived the same way from the $\mathcal{D}_\mathrm{KL}$, with the difference being the inclusion of the noise $\xi_N$ as a parameter in the FisherNet's derivation.
In the end result, this leads to the prefactor $t(\hat{\xi}_N)^{-1}$ in the term responsible for minimizing the reconstruction loss as well as the two additional terms
$\hat{\xi}_N^2$and $\tr\left( \ln(t(\hat{\xi}_N))\right)$, which are not included in the ELBO.) 

\subsection{Sampling from the Inverse Metric} \label{sampling_chapter}
In the latent space of the FisherNet, we need to sample the latent space variables from the approximate posterior distribution $Q(z_i|d_i)=\mathcal{G}(z_i-\mu_i,M(\mu_i)^{-1})$, with the inverse metric as the covariance.
The covariance always has the form

\begin{align}
M(\mu_i)^{-1} &= \left( J(\mu_i)^{\dagger}I_d(f(\mu_i))J(\mu_i) + \mathds{1}\right)^{-1} \\ 
&= \qty(J(\mu_i)^{\dagger}N^{-1}J(\mu_i) + \mathds{1})^{-1}. 
\end{align}
In order to draw samples from this, we start by drawing samples from the parts of $M(\mu_i)$

\begin{align}
n^* &\sim \mathcal{G}\qty(n, N^{-1}) \\\
\eta^* &\sim \mathcal{G}(\eta,\mathds{1}).
\end{align}
We can combine the two partial samples via 

\begin{equation}
\vartheta^* = J(\mu_i)^{\dagger}n^* + \eta^* \label{eq:inv_sample}
\end{equation}
to obtain samples from the distribution given the inverse covariance.
\par
The next step is to apply the covariance to the sample of the inverse we have just drawn.
We can achieve this without explicitly calculating and inverting the metric by using the conjugate gradient algorithm~\cite{Shewchuk.1994} to solve 

\begin{equation}
\vartheta^* = M(\mu_i) z^* \label{eq:CG_setup}
\end{equation}
for $z^*$.
It can be verified by direct calculation that $z^*$ generated this way is drawn from a Gaussian with mean $\mu_i$ and covariance $M_i^{-1}$.
More details on the sampling process are provided in~\cite{Knollmuller.2019}.
\par
This sampling equation from two simple random variables to the latent space samples is still differentiable for the variational parameters—in our case, the mean $\mu_i$.
This access to the gradients of the variational parameters allows us to use the reparametrization trick~\cite{Kingma.2013} to train the FisherNet.

\section{Results}
In order to evaluate the performance of the FisherNet, we used the Fashion-MNIST dataset~\cite{Xiao.2017}.
The data set consists of $28$ by $28$ greyscale images of clothing items categorized into ten classes with  60,000 training samples and 10,000 test samples.
For comparison purposes, we also trained a VAE and a convolutional VAE~(CVAE).
We chose the hyperparameter configurations of the models to make a direct comparison, by using the same parameters whenever possible and using the same number of neurons in the hidden layers.
We did not perform extensive hyperparameter tuning to obtain optimal results with the data set.
The exact hyperparameter configurations can be found in Appendix~\ref{app:hyperparam}.

\subsection{Reconstruction Error}
An intuitive point to start the evaluation of an autoencoder model is the reconstruction error.
Architecturally, the models are trained on reconstructing data that they are fed as inputs.
Furthermore, both the ELBO and the KL divergence in Equation~\eqref{eq:loss_function} include a term that we can interpret as the reconstruction loss and which stems from the likelihood.

\subsubsection{Loss Function Behavior Versus Reconstruction Loss}
The reconstruction loss is just one term in the loss function of the FisherNet. Therefore, we first want to check how the reconstruction loss behaves compared to the actual loss function.
The loss function of the FisherNet is given by the KL divergence in Equation~\eqref{eq:loss_function}.
To see how well a FisherNet trained with this loss function delivers good reconstructions, we evaluated the loss function and the reconstruction loss for different latent dimensions on test data.
\par
To compare the models, we define the mean squared error (MSE) as a measure of the reconstruction error,

\begin{align}
\text{MSE}_i &= \frac{1}{k} \sum_{j=1}^k \left((d_j)_i - (\tilde{d}_j)_i\right)^2 \\
\text{MSE} &= \frac{1}{p} \sum_{i=1}^{p} \mathrm{MSE}_i \, ,
\end{align}
where the index $i$ identifies the data vector and $j$ specifies the vector component (or image pixel).
We average over the square of the difference between the $k$-dimensional data vector $d_i$ and the reconstructed data vector $\tilde{d}_i$.
To obtain a single value per model and training iteration, we then averaged this $\text{MSE}_i$ over all $p$ data points.
\par
\begin{figure}[H]
      \includegraphics[width=0.8\linewidth]{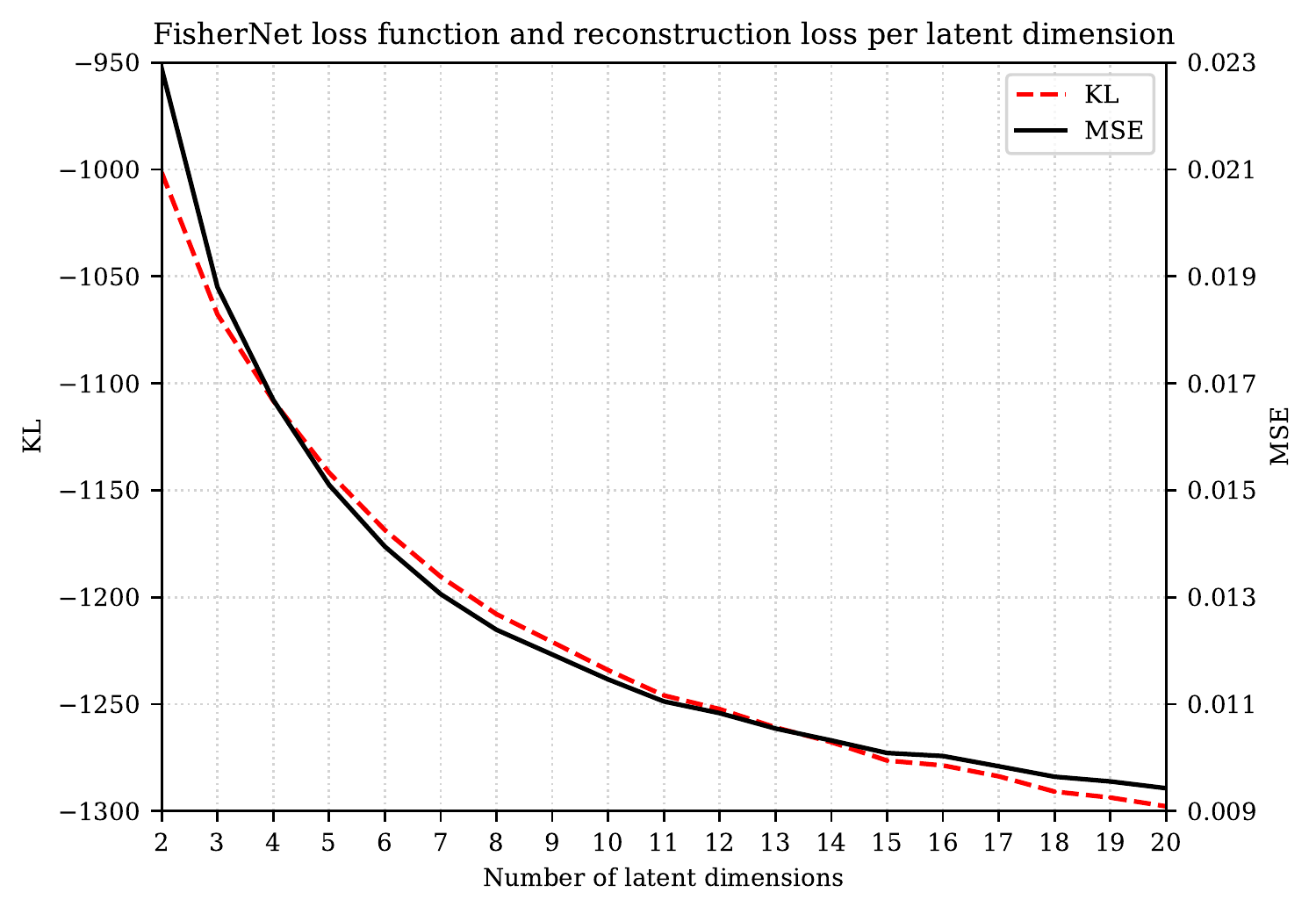}
  \caption{Variation of the loss function of the FisherNet and the reconstruction loss for different latent dimensions calculated for the Fashion-MNIST test data set. We did not include the constant terms of the KL divergence since they have no effect on the minimization. This explains why we obtain negative KL values.}
\label{fig:MSE_vs_loss}
\end{figure}

In Figure \ref{fig:MSE_vs_loss}, we show both the loss function and the reconstruction error on test data against the number of latent dimensions.
We can see that training the FisherNet by minimizing the KL simultaneously minimizes the reconstruction loss by an amount roughly proportional to the KL decrement for the different number of latent dimensions.
\par
\begin{figure}[H]
      \includegraphics[width=0.8\linewidth]{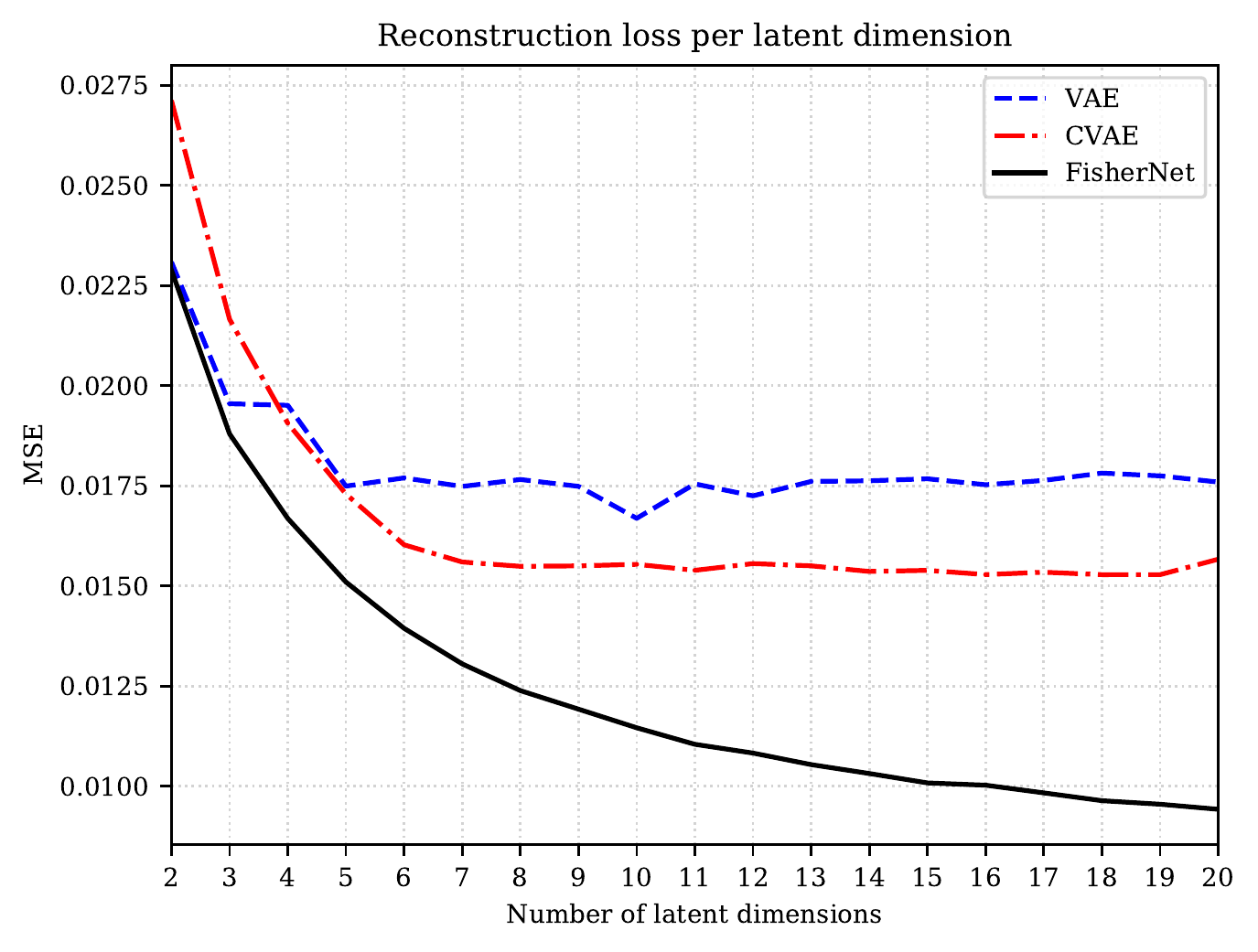}
  \caption{Comparison of the MSE on test data of the Fashion-MNIST data set for different number of latent dimensions.}
\label{fig:MSE_per_LD}
\end{figure}

In addition, as can be seen in Figure~\ref{fig:MSE_per_LD}, the reconstruction loss and KL divergence improve rapidly with additional latent space dimensions for smaller numbers of dimensions, but this trend slows down without stalling for all 20 latent dimensions we displayed here.
We expect the loss function to stagnate at a fixed value.
\par
Our interpretation of the stagnating loss function and reconstruction loss is that the model architecture can only gain a certain amount of non-redundant information from the data.
Therefore, the loss function improves very rapidly in the beginning, when the FisherNet can use each additional latent dimension to encode important, non-redundant information.
When most of this information is used by the FisherNet, the improvement slows down. The FisherNet can only use the additional latent space dimensions for largely redundant information.

\subsubsection{Model Comparison}

We have now seen that the FisherNet is well trained to minimize the reconstruction loss, so we want to compare it to the performance of the VAE and the CVAE.
To best compare the different architectures, we gave the FisherNet and the VAE equivalent architecture layouts.
The CVAE's number of weights differs from the other two networks as it is a convolutional network, and the VAE and FisherNet are fully connected networks.
\par
We repeatedly trained all three models on the Fashion-MNIST data set, starting with a latent space with two dimensions, which we subsequently increase.
Figure \ref{fig:MSE_per_LD} displays the MSE on the test data as a function of the number of latent dimensions for all three networks.

\par
The number of layers and nodes of the CVAE and the VAE agree, but their connections differ.
A VAE is fully connected and therefore has all the local connections of the CVAE and additionally all non-local connections.
In principle, the VAE architecture allows to fully emulate a CVAE, as a convolutional network  can  completely be represented by a fully connected network.
Since the VAE has more freedom than the CVAE, it should in principle always perform better than the latter, if perfectly trained.
\par
However, in practice, training a VAE can be more difficult than training a  CVAE. This is because if translational invariance of the features is a principle that is supported  by the data, the VAE itself has to discover this during training, whereas this information is already coded into the architecture of the CVAE.
\par
Thus, under limited training, limited either because of finite training time or finite training data, the CVAE can be more efficient than the VAE. This is reflected in \mbox{Figure \ref{fig:MSE_per_LD}}, where the CVAE's reconstruction loss starts much higher than the VAE's for low-dimensional latent spaces, where the VAE can be optimized fully.
For the larger latent spaces, however, the VAE is clearly not fully optimized.
It saturates early and fluctuates around an MSE value of approximately $0.0175$.
The CVAE, on the other hand, outperforms the VAE for the higher-dimensional latent spaces, showing how the constraints on the convolutional layers lead to an advantage when it comes to training efficiency.
\par
The FisherNet is a fully connected network that agrees with the VAE in the number of layers and nodes.
It differs from the VAE by the approach to the inference of the latent space variables.
For the FisherNet, we reduced the number of variational parameters to only the mean and approximate the uncertainty with the metric in Equation~\eqref{eq:Gaussian_fishermetric}.
This allows the FisherNet to use correlations in latent space.
This change affects the reconstruction loss in two ways.
First, the FisherNet gains an advantage through the exploitation of the latent space correlations.
This advantage is relatively small but scales with the number of latent dimensions.
The second effect is that the need to learn variances is alleviated, which leads to a more straightforward training process, because inferring them jointly with the means is known to be more difficult than learning the means alone~\cite{Bishop.2006}.
\par
The first of those effects can be seen for the two and three-dimensional latent spaces in Figure \ref{fig:MSE_per_LD}.
The FisherNet's reconstruction error starts slightly below the VAE's at two latent dimensions.
This difference in the MSE value increases in the step to three latent dimensions from an MSE value of $2\times 10^{-4}$ to $7.5\times 10^{-4}$ at three latent dimensions.
For larger latent spaces, this effect is no longer visible since the VAE's reconstruction error starts to fluctuate when the VAE is no longer fully optimized.
\par
This leads us to the second effect. 
The FisherNet is more straightforward to train, which means it can utilize the theoretical advantage it has over the CVAE.
The FisherNet keeps steadily improving its reconstruction error.
Unlike the VAE, the FisherNet avoids the training issues of not fully converging, and gains a significant advantage, which grows with the number of latent dimensions.

\subsubsection{Training Performance}

\begin{figure}[H]
      \includegraphics[width=\linewidth]{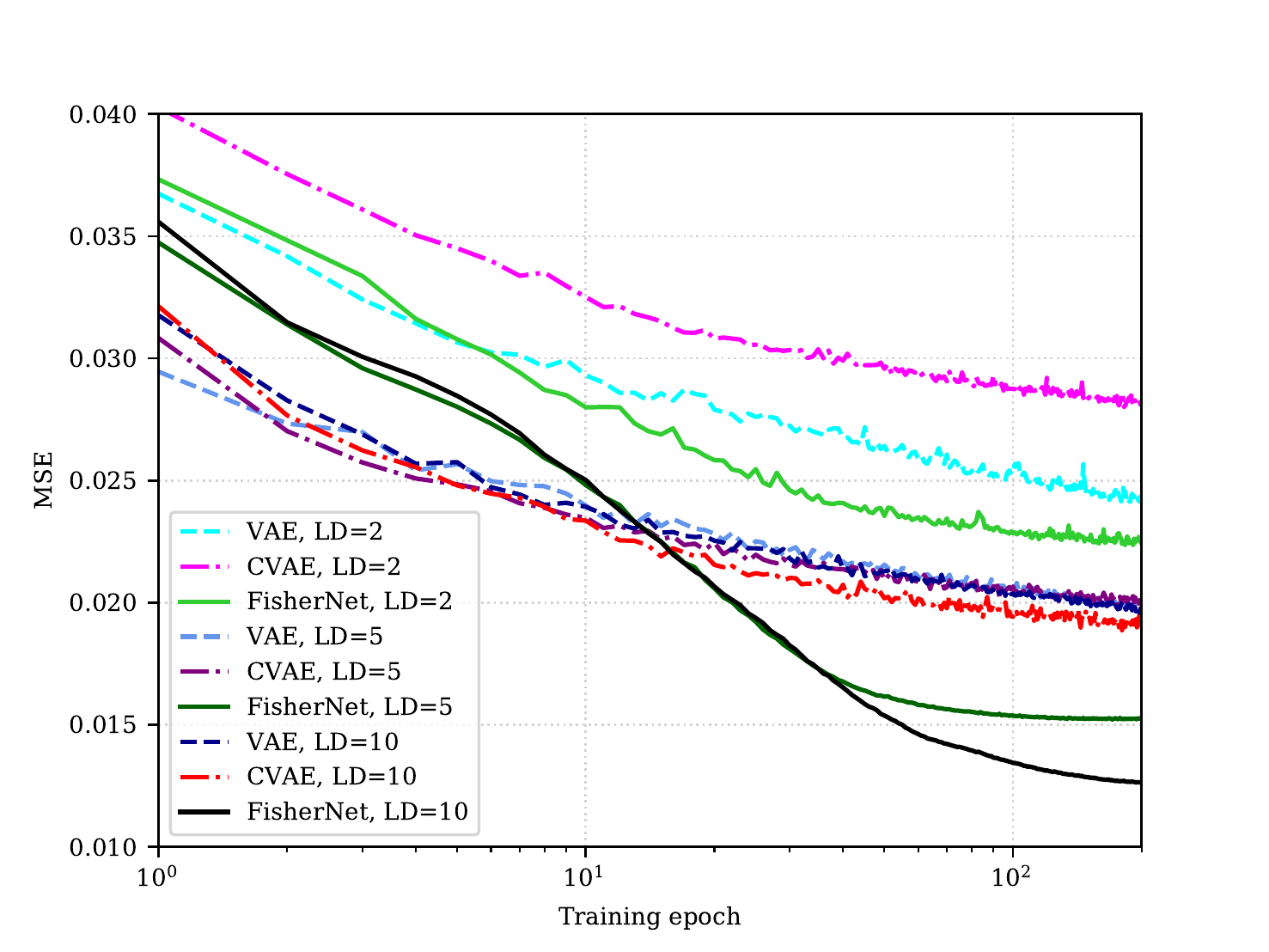}
  \caption{Comparison of the MSE on test data of the Fashion-MNIST data set after each of the first 50~epochs. This comparison is shown for the FisherNet, the VAE, and the CVAE for 2, 5, and 10 latent dimensions (LD).}
\label{fig:MSE_per_epoch}
  \end{figure}

  Analyzing the reconstruction error for different latent dimensions shows that the FisherNet avoids optimization problems the VAE runs into for a high number of latent dimensions.
To gain more insight into this, we now look at the behavior of the reconstruction loss during training.
Figure \ref{fig:MSE_per_epoch} displays the behavior of the reconstruction loss during optimization for the three models using 2, 5, and 10-dimensional latent spaces.

\par
As we have seen before, the FisherNet has fewer variational parameters than the VAE since we eliminated the latent space covariance as a variational parameter.
Instead, we approximate it with a metric that is informed directly by the generative process.
Therefore, the initial optimization of the FisherNet is slower than for the other two models.
These effects are clearly shown in Figure \ref{fig:MSE_per_epoch}.
The FisherNet's reconstruction loss starts higher than the other two models after the first training epoch.
This behavior is reproduced by all of the evaluated latent space dimensionalities.
During the next few epochs, the improvements are also slower than for the other models.
\par
However, the reduction in variational parameters and the fact that the FisherNet's latent space distribution is informed directly from the generative process leads to a more steady optimization.
Therefore, while the FisherNet's optimization starts slow, it outperforms the other two models given a long enough training time.
This is visible in Figure \ref{fig:MSE_per_epoch}, where the reconstruction loss of the FisherNet using a two-dimensional latent space is lower than both the corresponding losses of the other two models after twelve epochs.
If we look at the higher dimensional models, this becomes even more apparent.
The FisherNet using a five-dimensional latent space outperforms both of the other models with ten latent dimensions after forty epochs.
\par

\subsection{Analyzing the Latent Space}
To perform variational inference for the latent space variables, we demanded the latent space distribution to be continuous.
A result of this continuous distribution is that the encoder infers nearby mean positions for similar data points.
This property of the latent space opens the door for methods such as representation learning~\cite{Bengio.2012, Tschannen.2018}.
Representation learning attempts to find useful information about a data set in the latent space representation.
This can be useful for many different problems since its latent space usually has much lower dimensionality than the input data.
Representation learning is often used as a preprocessing step before further analyzing a data set.
\par
\begin{figure}[H]
     \includegraphics[width=\linewidth]{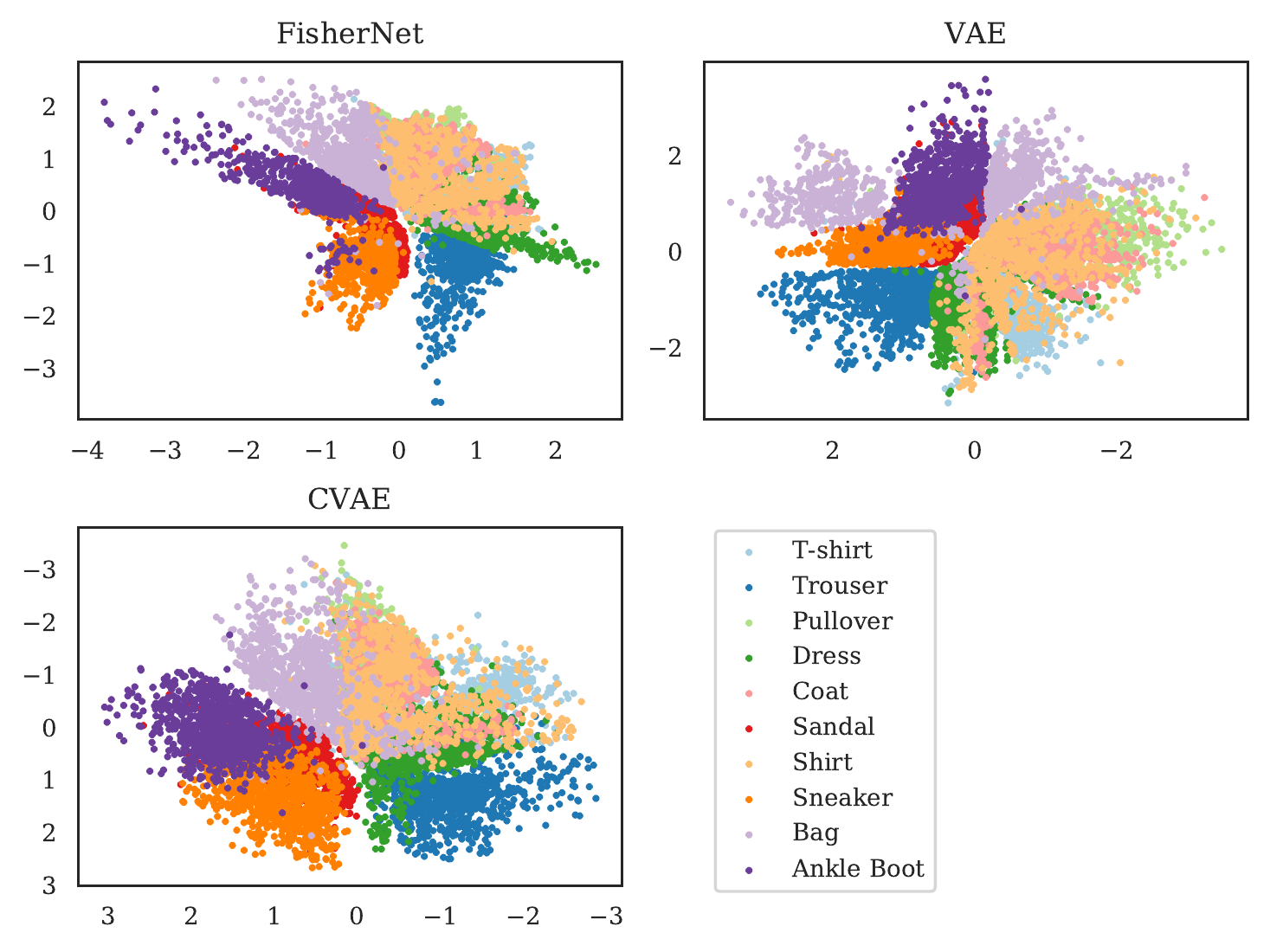}
\caption{Two-dimensional latent space means $\mu_i$ for all $10^3$ images in the Fashion-MNIST test data set. The top left shows the inferred means for the FisherNet, the top right for the VAE, and the bottom left for the CVAE. The means $\mu_i$ are color-coded by their category provided with the Fashion-MNIST data set.
For improving the visual similarity of these distributions, the VAE distribution was mirrored at the $x$-axis and the CVAE at the $x$- and $y$-axis.}
\label{fig:LS_comp}
 \end{figure}

We showed above that the FisherNet has an improved reconstruction accuracy compared to the standard VAE and converges better for high dimensional latent spaces.
Next, we look at the latent space representations to see if the FisherNet preserves the beneficial latent space qualities VAE's bring.
To start the analysis of the latent space, we use two-dimensional latent spaces.
For this purpose, we calculated all mean positions $\mu_i$ for the images in the test data using the encoders of the respective models.
We made a scatter plot of all these means $\mu_i$ in Figure \ref{fig:LS_comp}, where we color-coded the means by the classes of the images, from which these encoders inferred the mean positions.
This reduces the dimensionality of the data from 784, the number of pixels in the input images, to 2.

  \par
Since all three of the models we analyze here rely on variational inference for the latent space variables, the general structure of how the means of the different classes are ordered is similar.
The latent space positions, as inferred from the data, are close to each other for similar images.
This leads to a clustering, with the images belonging to a class  grouped together.
Since the similarity is given at the edges of a cluster as well, similar classes are grouped together as well, while classes that are visually distinct from each other are separated.
\par
Figure \ref{fig:LS_comp} displays these effects for the latent spaces of the three models we are analyzing.
For example, we can see that the FisherNet and the CVAE group together the three classes representing shoes.
In contrast, they group the clothing articles flanked by trousers on one side and bags on the other.
These groupings line up with a human evaluation of the similarity between the images.
The VAE generally shows similar groupings but puts the mean positions of some of the bags into the shoe cluster.
As we expected, this result shows us that the FisherNet preserves the grouping tendencies of the VAE models in latent space.
   {We analyzed the grouping tendencies further in Appendix \ref{app:clustering} with the use of the k-means algorithm.
}  

\subsubsection{Uncertainties}

Since we are approximating the distribution for a point in latent space corresponding to a data point with a Gaussian, we have the mean $\mu_i$ and the covariance $M(\mu_i)^{-1}$ from Equation~\eqref{eq:Gaussian_fishermetric}.
To analyze the uncertainties around a mean, we can perform an eigenvalue decomposition of the covariance matrix or, equivalently, of its inverse $M(\mu_i)$.
This provides us with the directions and magnitudes of the uncertainties in terms of the eigendirections and the inverse square roots of the eigenvalues of $M$, respectively.
The VAE's encoder infers the uncertainty directly from the data.
Since this happens utilizing a diagonal covariance, the uncertainty direction of the inferred standard deviation corresponds to the axis of the latent space.
\par
Since the FisherNet's inference allows correlations and the uncertainty can have any direction  in the  latent space, the FisherNet can approximate the local latent space posterior more closely.
This combined with the uncertainty metric being a lower bound to the true posterior uncertainty leads to a smaller uncertainty area than the one inferred by the VAE. 
\par
\begin{figure}[H]
      \begin{subfigure}[b]{0.49\linewidth}
\includegraphics[width=\linewidth]{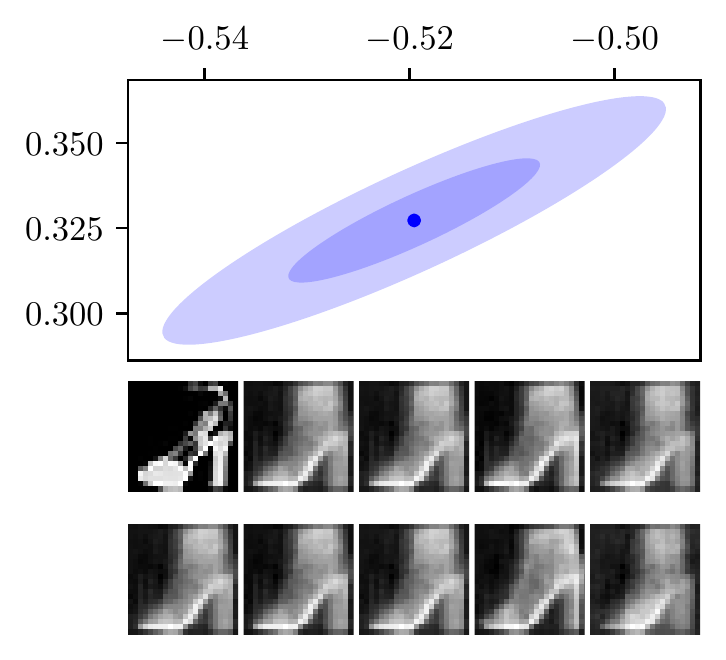}
\caption{\centering}
         \label{fig:fisheruncertsamples}
     \end{subfigure}
     \begin{subfigure}[b]{0.49\linewidth}
         \includegraphics[width=\linewidth]{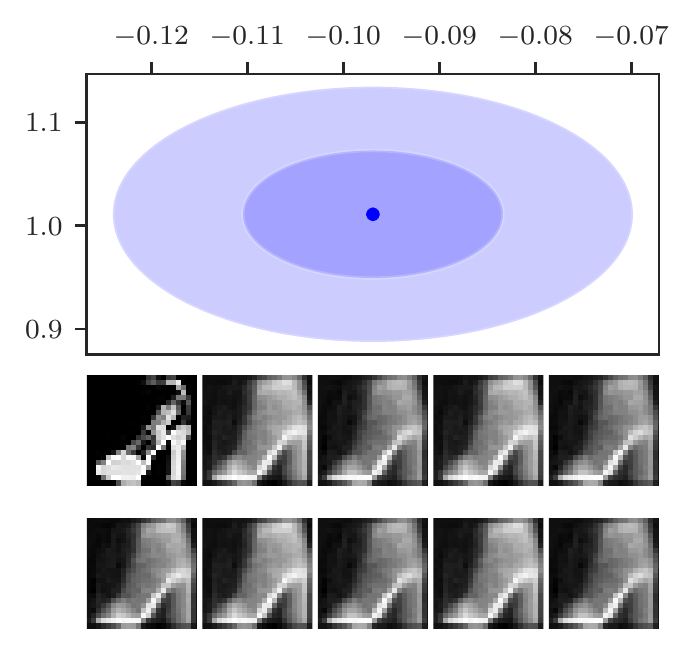}
         \caption{\centering}
         \label{fig:vaeuncertsamples}
     \end{subfigure}
\caption{(\textbf{a}) {FisherNet}. (\textbf{b}) VAE. The top shows the one-$\sigma$ and two-$\sigma$ uncertainty areas around the latent space mean inferred from a data point for (\textbf{a}) the FisherNet (left) and (\textbf{b}) the VAE (right), both using two-dimensional latent spaces. For the FisherNet, we calculated $\sigma$ from the eigenvalues of the uncertainty metric. Below the uncertainty plot, the very left image in the top row shows the original image. The very left picture in the bottom row shows the reconstruction from the mean position. The following four images are reconstructions from points at a distance of one-$\sigma$ in the directions of the uncertainty metrics eigenvectors in the top row. The four images in the bottom row are reconstructed from the points in 2-$\sigma$ distance from the mean. On the right, we applied the same method for the VAE using the inferred $\sigma$.}
\label{fig:Fishernet_uncert_samples}
\end{figure}

This is illustrated in Figure \ref{fig:Fishernet_uncert_samples}a, which shows the one and two-$\sigma$ uncertainties around the mean of a sample image from the Fashion-MNIST data set.
For comparison, the uncertainty area for the same sample as inferred by the VAE is displayed in Figure \ref{fig:Fishernet_uncert_samples}b.
We see that the off-kilter orientation of the largest uncertainty leads to a narrower uncertainty area.
\par
 Both models infer a continuous distribution over the whole latent space, where the mean positions of similar data points are close to each other.
Therefore, sampling around the mean inside the area provided by the uncertainty gives similar reconstructions to the reconstruction at the mean  position.
To illustrate this, Figure \ref{fig:Fishernet_uncert_samples}a also displayed the reconstructions by the FisherNet from the mean position as well as from points in one and two-$\sigma$ distance from the mean in both the positive and negative direction of the eigenvectors.
Figure \ref{fig:Fishernet_uncert_samples}b displays the mean and uncertainty for the same sample inferred by the VAE, as well as its corresponding reconstructions.
\par

  If we zoom out a little from a single point in latent space and instead focus on the cluster of the images representing different types of shoes, we can analyze more broad trends.
In representation learning, the independence of the latent space dimensions is often used as a way to force the model to use them to encode independent features of the data.
This can be a very useful preprocessing step before the data are analyzed further.
\par
The FisherNet, in contrast to this, allows for correlations between the latent space dimensions.
It, therefore, does not infer independent features.
Since the FisherNet's uncertainty approximation is based on the Fisher metric, which measures the effect of small variations, the direction of largest uncertainty should be towards the region least well determined by the data.
This allows the FisherNet to distinguish more clearly between different data points and leads to less overlap between the uncertainty areas of different points.
\par
\begin{figure}[H]
     \includegraphics[width=\linewidth]{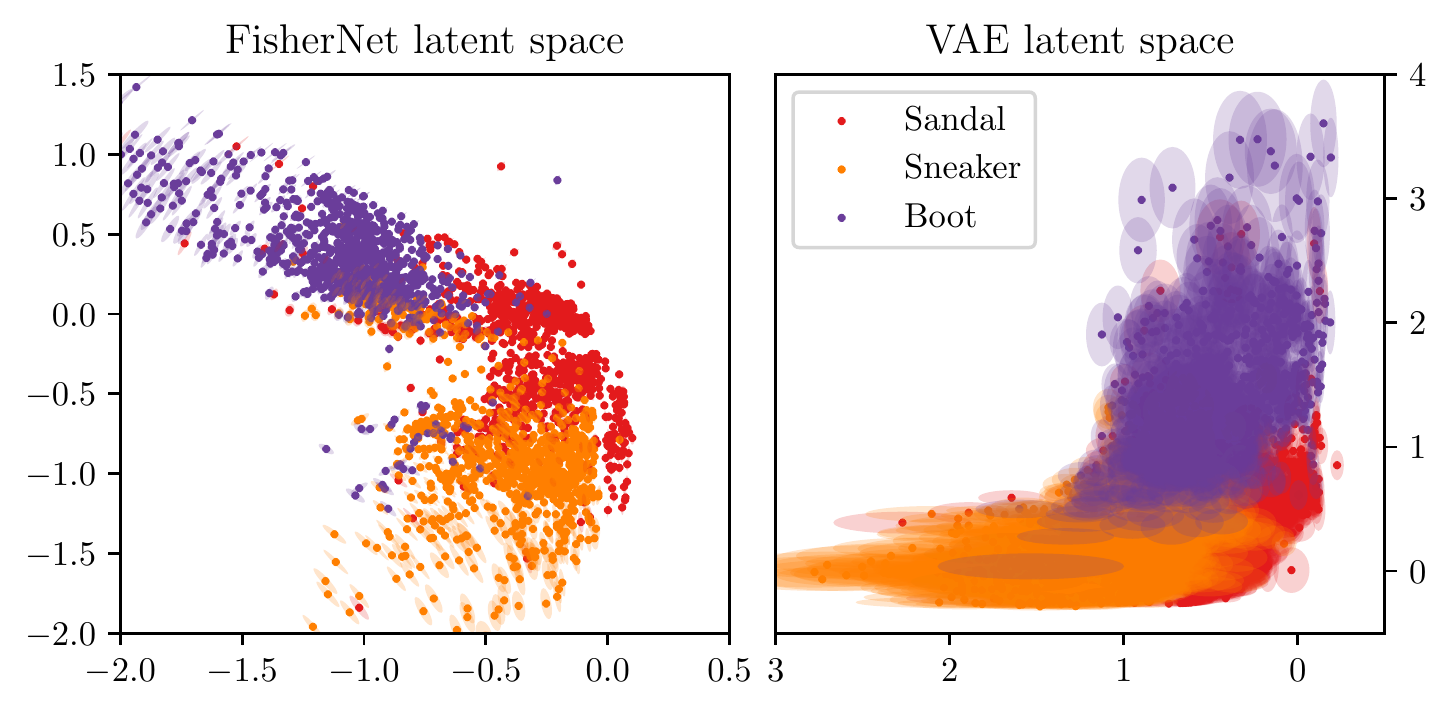}
\caption{The latent space means $\mu_i$ and their two-$\sigma$ uncertainty areas for the images belonging to the ``shoe'' classes of the Fashion-MNIST test data set. The left plot shows this for the FisherNet, and the plot on the right for the VAE. The legend shows the color-coding of the means and their two-$\sigma$ uncertainty areas for both plots.
The VAE distribution was mirrored at the $x$-axis.}
\label{fig:2sigma_areas}
 \end{figure}

Figure \ref{fig:2sigma_areas} shows the mean positions and the two-$\sigma$ uncertainty areas around them, within two-dimensional latent spaces, for these data points as found by the FisherNet and the VAE.
It illustrates how the need for independent features affects the inference of mean and uncertainty in the VAE.
The clustering of the means and the uncertainty areas occur in alignment with the latent space axis.
\par
The figure also illustrates that the uncertainty of the FisherNet is affected by how well the latent space locations are determined by data.
The uncertainties are very small in the region close to the origin, which is the most densely populated region.
The same goes for the region where the shoe cluster neighbors the bags as we saw in Figure \ref{fig:LS_comp}.
The uncertainties become larger for more peripheral points in the less dense regions.
The direction of the biggest uncertainty of the FisherNet is generally roughly orthogonal to the larger structure determined by the mean positions and therefore towards the less densely populated part of the latent space.
\par
We saw before that the latent space uncertainties are smaller and more narrow for the FisherNet than for the VAE.
This leads to less overlap between the uncertainty areas of the FisherNet's latent space, especially in less populated regions.

\subsubsection{Using the Metric}
We have seen that the FisherNet is not suited for finding independent features of the data in the latent space representation.
However, the FisherNet architecture gives us a local metric in the latent space that provides us with the approximate uncertainty covariance.
We anticipate that this metric can be used for analysis of the latent space representation.
We will, however, leave the exact uses of the metric up for future research.

\subsection{Generating Samples} \label{sec:generating_samples}
The next thing we want to look at, when it comes to evaluating the performance of the FisherNet compared to the more traditional VAE, is the quality of newly generated samples.
We use the Fr\'echet inception distance~(FID)~\cite{Heusel.2017} as the measure of sample quality.
The FID is a heuristic measure to evaluate the quality of generated images.
It is calculated as the Wasserstein metric between two multivariate Gaussian distributions parametrized by features, which are found via an Inception v3 network~\cite{Szegedy.2016} trained on ImageNet~\cite{Deng.2009}.
A small FID indicates that the generated images are similar to the real images.

\subsubsection{Generating Samples from a Unit Gaussian Distribution}

The most common way to generate new latent space samples of a VAE model is from a standard Gaussian distribution. 

Deriving the loss function via the KL results in a  regularizing term, which constrains the latent space posterior to stay close to the prior distribution in both the VAE's loss function, the ELBO 
\cite{Kingma.2013}, and the FisherNets loss function.
Since this prior is usually chosen to be a standard Gaussian distribution, this is an intuitive approach to take and for the VAE a sensible approach.
However, a known issue of this approach  is that the distribution of the latent space data representation chosen by VAEs does not always stay close to that of the assumed prior 
\cite{Kingma.2019}.
\par
We also chose a standard Gausssian prior for the latent space variables when we derived the FisherNet.
Therefore, with perfect training of the network, we should be able to find the generating function from this distribution to the data distribution.
Perfect training, however, is not achieved with the current training methods and the FisherNet's latent space distribution, especially in higher dimensions{, }  does not stay close to the postulated prior.
   {The very same problem is faced by  VAE as well, just slightly less severe there. 
}  Therefore, drawing standard Gaussian samples in the latent space is not generally a viable strategy to generate new images with the FisherNet.
\par
\begin{figure}[H]
         \begin{subfigure}[b]{0.32\linewidth}
\includegraphics[width=\linewidth]{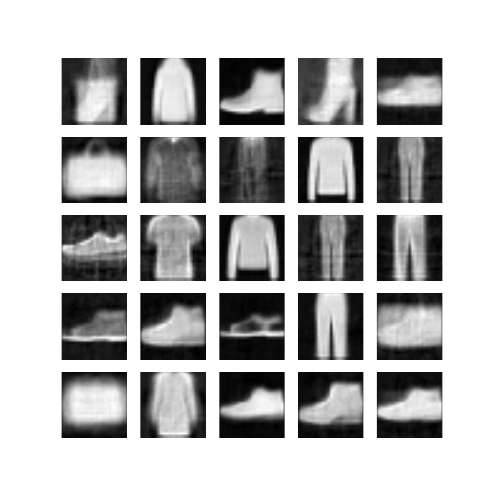}
\caption{\centering}
         \label{fig:fisher2dsamples}
     \end{subfigure}
     \begin{subfigure}[b]{0.32\linewidth}
         \includegraphics[width=\linewidth]{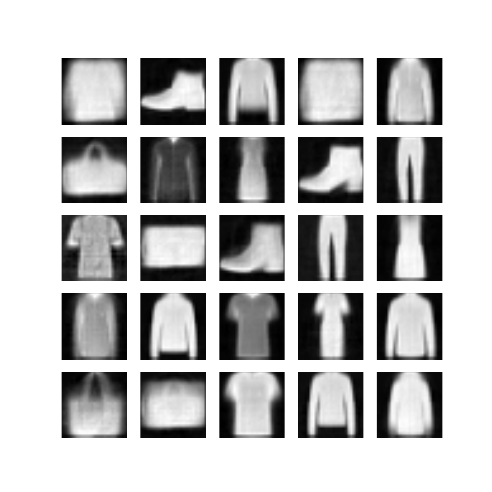}
         \caption{\centering}
         \label{fig:vae2dsamples}
     \end{subfigure}
     \begin{subfigure}[b]{0.32\linewidth}
         \includegraphics[width=\linewidth]{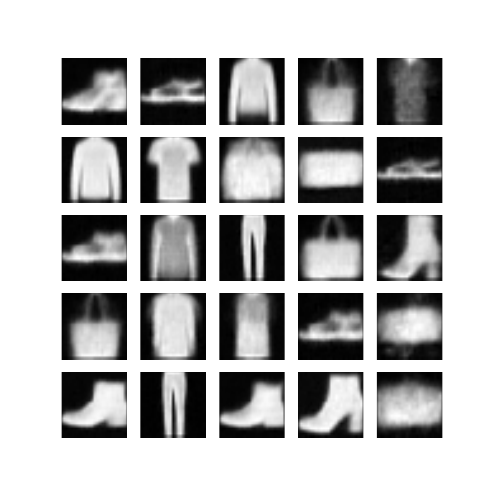}
         \caption{\centering}
         \label{fig:cvae2dsamples}
     \end{subfigure}
\caption{(\textbf{a}) FisherNet. (\textbf{b}) VAE. (\textbf{c}) CVAE. Newly generated samples of the Fashion-MNIST data set. Left using the FisherNet, middle the VAE, and right the CVAE. All three models used a two-dimensional latent space.}
\label{fig:2dsamples}
\end{figure}
\begin{figure}[H]
      \begin{subfigure}[b]{0.32\linewidth}
\includegraphics[width=\linewidth]{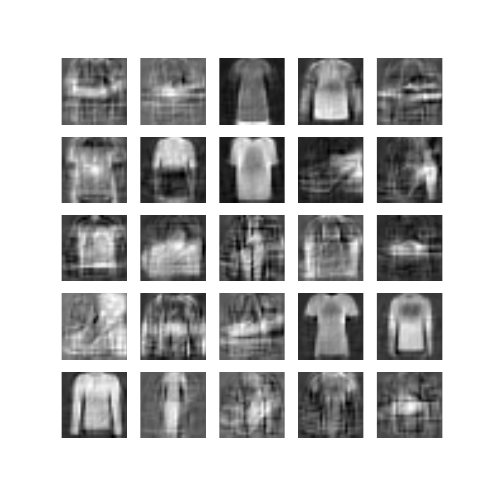}
\caption{\centering}
         \label{fig:fisher15dsamples}
     \end{subfigure}
     \begin{subfigure}[b]{0.32\linewidth}
         \includegraphics[width=\linewidth]{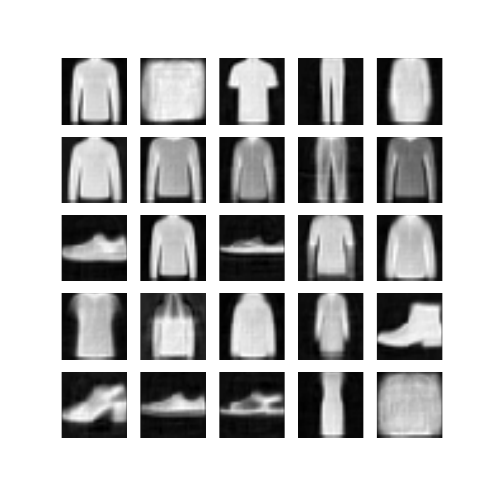}
         \caption{\centering}
         \label{fig:vae15dsamples}
     \end{subfigure}
     \begin{subfigure}[b]{0.32\linewidth}
         \includegraphics[width=\linewidth]{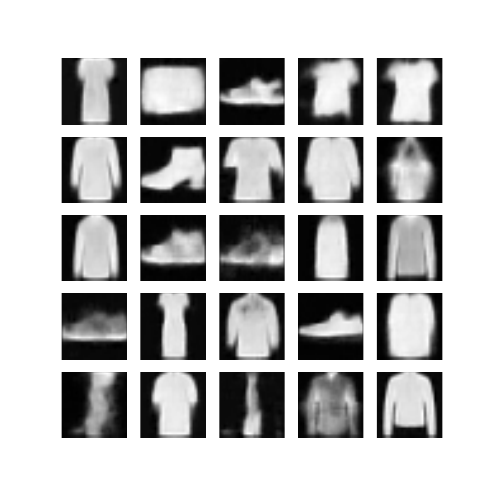}
         \caption{\centering}
         \label{fig:cvae15dsamples}
     \end{subfigure}
\caption{(\textbf{a}) FisherNet. (\textbf{b}) VAE. (\textbf{c}) CVAE. Same as Figure \ref{fig:2dsamples}, but using a 15-dimensional latent space.}
\label{fig:15dsamples}
\end{figure}

To demonstrate this, we drew $10^4$ samples from a unit Gaussian distribution and reconstructed the samples with the decoders of the three models we have been analyzing.
Figures \ref{fig:2dsamples} and \ref{fig:15dsamples} display a few of those newly generated images for 2- and 15-dimensional latent spaces, respectively.
Then we calculated the FID between the $10^4$ newly generated images and the images in the Fashion-MNIST test data set.
We show the resulting FIDs along with the FID scores of the reconstructed test data in Table \ref{tab:FID_scores_rec_wihtenoise}.
The FIDs of the reconstructions are included as a floor value for the model.
The reconstruction FIDs confirm our findings for the reconstruction error.
    \begin{specialtable}[H]
    \tablesize{\small}
\caption{FIDs for the test data reconstructions of the three models (rec) and newly generated images from standard Gaussian latent space samples (sample). For the FisherNet, the FIDs for samples generated using Mat\'ern kernel density estimation (MKDE) are included.}
\setlength{\cellWidtha}{\columnwidth/7-2\tabcolsep+0.0in}
\setlength{\cellWidthb}{\columnwidth/7-2\tabcolsep+0.0in}
\setlength{\cellWidthc}{\columnwidth/7-2\tabcolsep-0.0in}
\setlength{\cellWidthd}{\columnwidth/7-2\tabcolsep-0.0in}
\setlength{\cellWidthe}{\columnwidth/7-2\tabcolsep-0.0in}
\setlength{\cellWidthf}{\columnwidth/7-2\tabcolsep-0.0in}
\setlength{\cellWidthg}{\columnwidth/7-2\tabcolsep-0in}
\scalebox{1}[1]{\begin{tabularx}{\columnwidth}{>{\PreserveBackslash\raggedright}m{\cellWidtha}>{\PreserveBackslash\centering}m{\cellWidthb}>{\PreserveBackslash\centering}m{\cellWidthc}>{\PreserveBackslash\centering}m{\cellWidthd}>{\PreserveBackslash\centering}m{\cellWidthe}>{\PreserveBackslash\centering}m{\cellWidthf}>{\PreserveBackslash\centering}m{\cellWidthg}}
\toprule
                       &    \textbf{Latent Dimension}       & \textbf{2}   & \textbf{5}   & \textbf{10}  & \textbf{15}  & \textbf{20}  \\ \midrule
  & $\text{FID}_{\text{rec}}$    & 116 & 80  & 68  & 61  & 57 \\ \cmidrule{2-7} 
                        {FisherNet}   & $\text{FID}_{\text{sample}}$ & 116 & 110 & 124 & 135 & 148
    \\ \cmidrule{2-7} 
                           & $\text{FID}_{\text{MKDE}}$ & 116 & 85 &  &  &   \\ \midrule
 & $\text{FID}_{\text{rec}}$     & 125 & 97  & 98  & 97  & 97  \\ \cmidrule{2-7} 
                        {VAE}    & $\text{FID}_{\text{sample}}$ & 125 & 99 & 100  & 102  & 100 
                           \\   \cmidrule{2-7} 
                           &  $\text{FID}_{\text{MKDE}}$  &  124 &  97 &  &  &  \\   \midrule
    & $\text{FID}_{\text{rec}}$   & 151 & 95 & 86  & 87  & 90  \\ \cmidrule{2-7} 
                   {CVAE}          & $\text{FID}_{\text{sample}}$ & 152 & 95 & 89 & 92  & 100
                           \\  \cmidrule{2-7} 
                           &  $\text{FID}_{\text{MKDE}}$ &  151 &  95 &  &  &   \\  \bottomrule
\end{tabularx}}
\label{tab:FID_scores_rec_wihtenoise}
\end{specialtable}
\par
The FIDs calculated for the newly generated samples start favorably for the FisherNet.
At a two-dimensional latent space, the FID for the FisherNet's generated samples are close to the reconstruction FID and therefore better than the new samples of the other two models.
Looking at the higher-dimensional latent spaces, however, the FID of the FisherNet's samples does not improve much and falls behind its reconstructions.
The sample FIDs of the other two models stay relatively close to their reconstruction FID, illustrating how the ELBO's regularization keeps their latent space distributions close to the prior.
\vspace{-6pt}

\subsubsection{Generating Samples Using Density Estimation}
The FisherNet does not reach its reconstruction FIDs for the higher number of latent dimensions.
We can trace this to a generative function in Equation~\eqref{data_generating_function}, which we implemented with the decoder network.
Perfectly trained, we would be able to find a representation of this function for which the latent space posterior would be the same as the prior.
Since training the network perfectly is not possible with the current methods, we have to find a different way to sample using the FisherNet.
This problem is not unique to the FisherNet but common in all kinds of VAE models.
\par
\begin{figure}[H]
     \includegraphics[width=\linewidth]{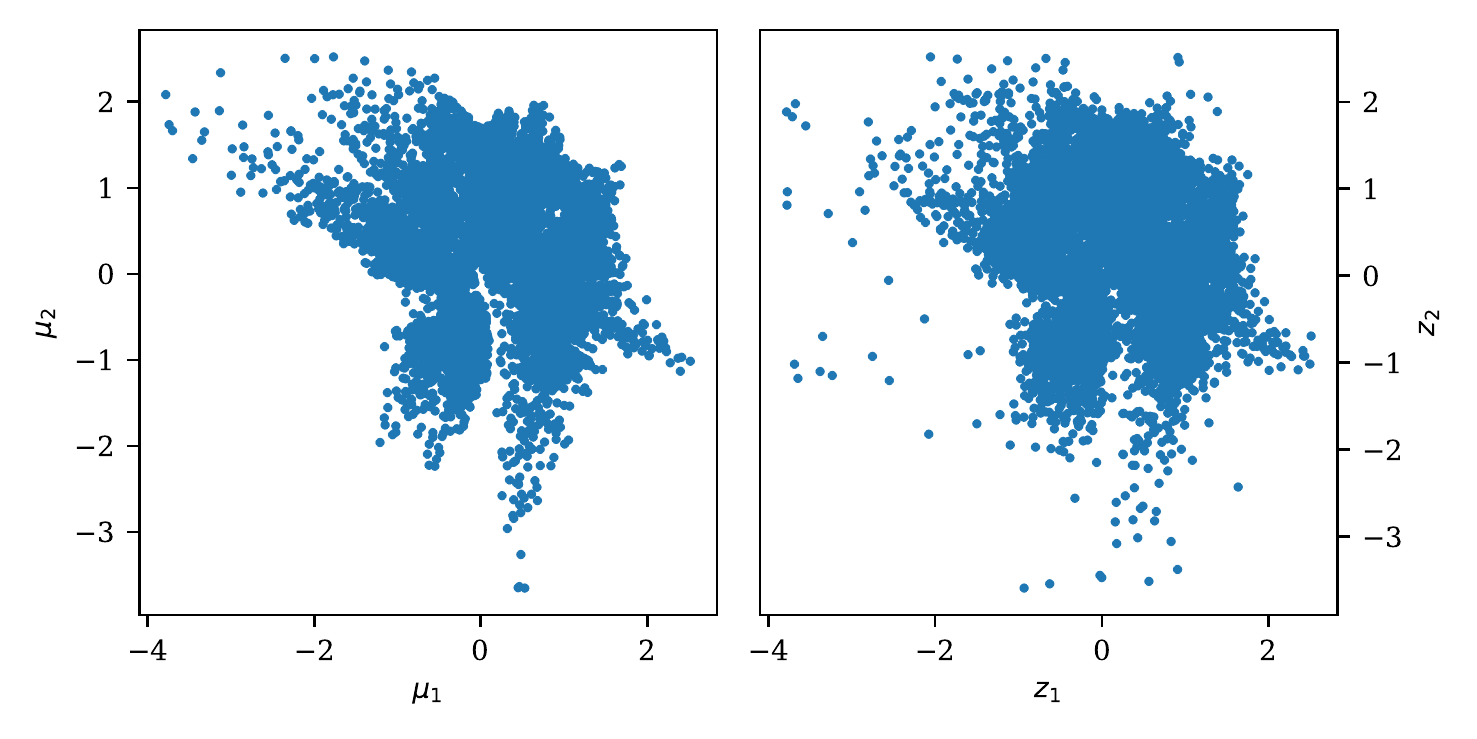}
\caption{Inferred mean positions of the test data in the two-dimensional latent space (\textbf{left}) and samples generated using MKDE (\textbf{right}).}
\label{fig:density_estimator_2dvis}
\end{figure}

A promising approach is to use a secondary density estimation in the latent space that allows us to introduce a new transformation $f'$ from a standard Gaussian distribution to the posterior distribution we found when training the network.
This lets us generate new data samples according to:
\begin{equation}
d_i^*  = f_\theta(z_i^*) = f_\theta \qty(f'(\eta^*)) ,
\end{equation}
starting from a standard Gaussian sample  $\eta^*$.
This method has been utilized in~\cite{Rezende.2015} using normalizing flows as the secondary density estimator.
To showcase this method combined with the FisherNet{, and for comparison the VAE and CVAE, }  we used Mat\'ern Kernel Density Estimation (MKDE)~\cite{Guardiani.2021} as the latent space density estimator, to generate the latent space samples.
MKDE is a fully Bayesian approach to reconstruct a smooth probability density from discrete data.
We performed MKDE on the FisherNet latent space distributions for latent spaces with two and five dimensions.
The samples drawn using the density estimation    {for the FisherNet latent space }  in two dimensions are displayed in Figure~\ref{fig:density_estimator_2dvis} alongside the latent space mean positions corresponding to the test data.
\par
\begin{figure}[H]
     \begin{subfigure}[b]{0.49\linewidth}
\includegraphics[width=\linewidth]{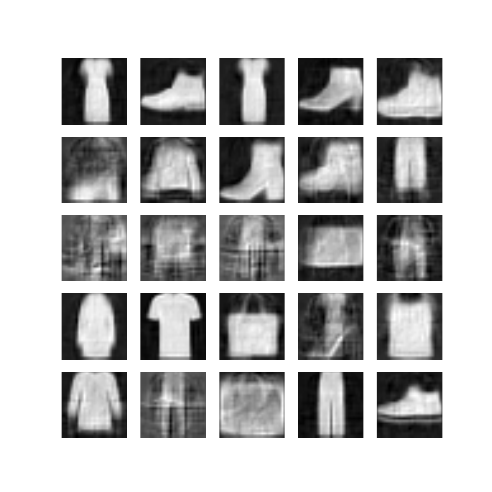}
         \label{fig:fisher15dsamples}
     \end{subfigure}
     \begin{subfigure}[b]{0.49\linewidth}
         \includegraphics[width=\linewidth]{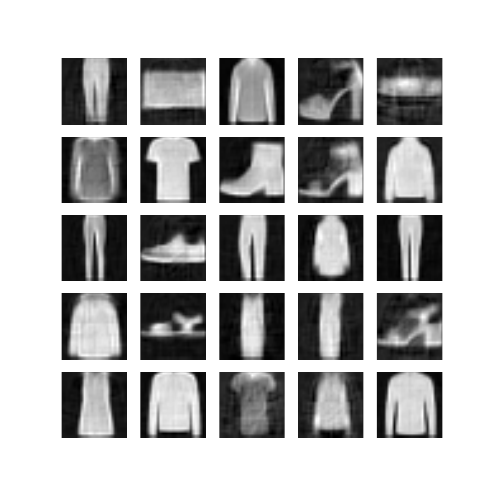}
         \label{fig:vae15dsamples}
     \end{subfigure}
\caption{Images generated by the FisherNet from a five-dimensional latent space using Gaussian samples (\textbf{left}) and samples generated via the density estimator (\textbf{right}).}
\label{fig:5den_dsamples}
\end{figure}

Using the density estimator to draw samples from the posterior latent space distribution improves the quality of the samples we can generate with the FisherNet.
For a two-dimensional latent space, the samples we generate this way match the FID score of 116 that we also found using standard Gaussian samples in Table~\ref{tab:FID_scores_rec_wihtenoise}.
For a five-dimensional latent space, the FID score is 85, which indicates a significant improvement over the results using standard Gaussian latent space samples and comes closer to the reconstruction score.
   {Using this method for the VAE and CVAE also slightly improves the sample quality, but since the reachable limit for sample quality is the reconstruction FID, this improvement is less notable.
}Therefore, we conclude that the FisherNet combined with a secondary density estimator for the latent space outperforms the VAE as a generative model.
Some sample images generated with this method can be seen in Figure~\ref{fig:5den_dsamples}.
This method of generating images is limited by the density estimation used in the latent space, since the common methods, including the one we used here, scale badly with the number of dimensions.
It also requires the training of a density estimator in addition to the FisherNet.

\section{Conclusions}

We derived a new variant of VAEs by expanding the scope of the variational inference beyond the mean-field approach.
We used the state-of-the-art inference technique of MGVI~\cite{Knollmuller.2019} and adapted its approach to the framework of deep generative models.
Analogous to~\cite{Milosevic.2021}, we derived the theoretical framework of the FisherNet as a modified VAE.
\par
While deriving the FisherNet, we made a number of simplifying assumptions:
The first simplification we made was assuming the noise to be Gaussian, isotropic, and the same everywhere in data space.
Next, we assumed a priori independence for the parameters $\theta$, $\xi_N$, and $Z$.
A priori independence is a prevalent assumption in Gaussian problems.
More drastically, we next assumed a posteriori independence for $\theta$ and $\xi_N$ in the approximating distribution $Q_\phi$.
This simplifying assumption allowed us to use MAP estimates for $\theta$ and $\xi_N$.
We made our final assumption in the calculation of the KL divergence when we assumed the metric approximating the covariance to be constant near the mean, at which we calculated it.
We list these simplifications here since they are informative about the limitations of our model and provide good starting points for possible improvements in further research.
\par
We evaluated the FisherNet on the Fashion-MNIST data set and compared its performance to a fully connected and a convolutional VAE.
We showed that the FisherNet's loss function correlates well with the reconstruction improvement of the input data.
We showed that the FisherNet outperforms the VAE in terms of reconstruction quality and scales better with a higher dimensionality of the space.
We showed that the reduction in variational parameters, when going from VAEs to FisherNets, by using an uncertainty approximation informed by the generative process in the latter models, leads to an initially slower but later saturated and improved optimization process.

\par
Our experiments showed that the FisherNet preserves the grouping tendencies of VAEs in the latent space distribution.
We showed how the FisherNet uses latent space correlations to achieve smaller uncertainties in the local posterior, allowing for a finer resolution between the different data points.
This is aided by the covariance approximation being a lower bound to the posterior uncertainty.
The correlations may make the FisherNet harder to use for some representation learning tasks such as component separation, since we do not impose a local posterior independence on the latent space variables.
This should, however, only have a small impact on the overall latent space distribution, since the VAE's independence constraints are also only local and both models start from a standard Gaussian prior for the latent space.
The availability of a local metric might open up new ways for latent space analysis.
We leave this option open for future research.
\par
Finally, we evaluated the FisherNet's performance on generating new artificial samples for the data set.
To this end, we first used standard Gaussian latent space samples to generate new images of the Fashion-MNIST data set and evaluated their quality by calculating the FID of the test data.
This experiment showed that the FisherNet's latent space distribution does not generally stay similar to the standard Gaussian prior.
Therefore, we concluded that drawing standard Gaussian samples is not viable for generating new data samples with the FisherNet.
However, we could demonstrate that the FisherNet can be a good model for generating new data samples when combined with an additional density estimator in latent space, such as an MKDE.
Using this approach, the higher reconstruction quality we achieved leads to a higher sample quality than we can achieve using the comparable VAE.
This approach is, however, limited since many density estimators, including MKDE, scale badly with the number of dimensions.
Overcoming these limitations is a challenge we hope will be solved by future research.
  \par
 {The FisherNet's current implementation is only a prototype and is unfortunately limited to small latent spaces. 
Therefore, we can not yet validate these promising results we achieved on Fashion-MNIST on more complex data.
We present the FisherNet architecture as a proof of concept for utilizing the generator model structure of the decoder for improving the variational inference of the posterior distribution.
}  

\vspace{6pt}
\authorcontributions{{Conceptualization, P.F.; Formal analysis, J.Z.; Software, J.Z.; Supervision, T.A.E.; Writing – original draft, J.Z.; Writing – review \& editing, P.F. and T.A.E. All authors have read and agreed to the published version of the manuscript.} 
}

\funding{This research received no external funding} 


\dataavailability{The data used in this article is the publicly available Fashion-MNIST data set available online at \url{https://github.com/zalandoresearch/fashion-mnist} 
} 


\conflictsofinterest{The authors declare no conflict of interest.}


\abbreviations{Abbreviations}{%
The following abbreviations are used in this manuscript:\\

\noindent 
\begin{tabular}{@{}ll}
AE & autoencoder \\
VAE & variational autoencoder\\
GAN & generative adversarial network\\
VI & variational inference\\
KL & Kullback--Leibler\\
MGVI & metric Gaussian variational inference\\
MAP & maximum a posteriori\\
ELBO & evidence lower bound \\
CVAE & convolutional variational autoencoder \\
MSE & mean squared error \\
FID & Fr\'echet inception distance
\end{tabular}}
\appendixtitles{no} 
\appendixstart
\appendix
\section{} \label{app:hyperparam}
We provide the full list of hyperparameters used to produce the results discussed in this paper in Table \ref{tab:hyperparameters}.
Please note that the implemented models include additional reshaping layers, as well as additional dense layers to go from the described network to the actual latent space representation and in the case of the CVAE from the latent space to the transposed convolutional decoder network.
\end{paracol}
\nointerlineskip
 \begin{specialtable}[H]\appendix
    \tablesize{\small}
\widetable
\caption{Hyperparameters.}
\setlength{\cellWidtha}{\columnwidth/4-2\tabcolsep+0.0in}
\setlength{\cellWidthb}{\columnwidth/4-2\tabcolsep+0.0in}
\setlength{\cellWidthc}{\columnwidth/4-2\tabcolsep+0.0in}
\setlength{\cellWidthd}{\columnwidth/4-2\tabcolsep+0.0in}
\scalebox{1}[1]{\begin{tabularx}{\columnwidth}{>{\PreserveBackslash\raggedright}m{\cellWidtha}>{\PreserveBackslash\centering}m{\cellWidthb}>{\PreserveBackslash\centering}m{\cellWidthc}>{\PreserveBackslash\centering}m{\cellWidthd}}
\toprule
\textbf{Hyperparameter} & \textbf{FisherNet} & \textbf{VAE} & \textbf{CVAE} \\ \midrule
Number of layers in de- and encoder & 3 & 3 & 3 \\
Neurons per layer & 448 & 448 & 448 \\
Layer type encoder & Dense & Dense & convolutional \\
Layer type decoder & Dense & Dense & transposed convolutional \\
Filters for convolutional layers & - & - & $\qty[32, 32, 64]$ \\
Filters for transposed convolutional layers & - & - & $\qty[64, 32, 32]$ \\
Kernel size & - & - & 3 \\
Strides & - & - & $[2,1,2]$ \\
Activation function & ReLU & ReLU & ReLU \\
Optimizer & Adam & Adam & Adam \\
Learning rate & $10^{-4}$ & $10^{-4}$ & $10^{-4}$ \\
Batchsize & 64 & 64 & 64 \\ \bottomrule
\end{tabularx}}
\label{tab:hyperparameters}
\end{specialtable}
\vspace{-12pt}
\begin{paracol}{2}
\switchcolumn
\appendix
  \section{} \label{app:clustering}

\textls[-15]{To analyze how well the latent space data representations belonging to a certain input data category are grouped together in the latent space, we apply the k-means algorithm to the latent space means of the test data as found by the encoders of the FisherNet, the VAE, and  the CVAE.
We then calculate for each data category the percentage of mean positions belonging to each cluster as found by the k-means clustering.
We then arrange this into a matrix in a way that maximizes the matrix trace by exchange of matrix columns. 
Figure~\ref{fig:5dcluster} shows a visualization of such matrices for the three architectures for a five-dimensional latent space.
The traces of these matrices are a measure of how well the k-means clusters fit the data labels and are provided in Table~\ref{tab:cluster}.
These results show that there is no significant difference between the way the three models group the data representations in latent space.}
 
\begin{figure}[H]\appendix

     \begin{subfigure}[b]{0.33\linewidth}
\includegraphics[width=\linewidth]{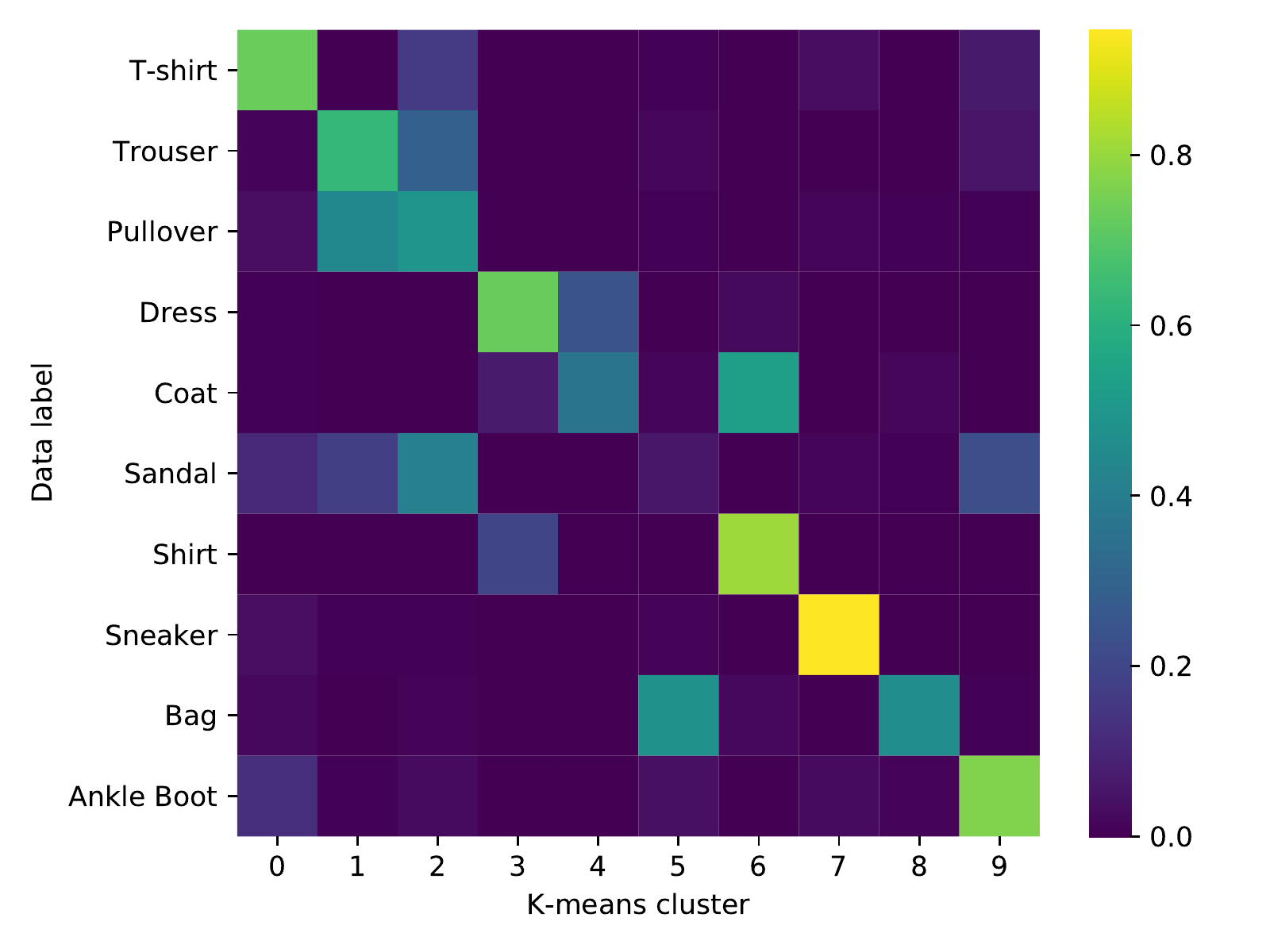}
\caption{\centering}
         \label{fig:fisher2dcluster}
     \end{subfigure}
     \begin{subfigure}[b]{0.33\linewidth}
         \includegraphics[width=\linewidth]{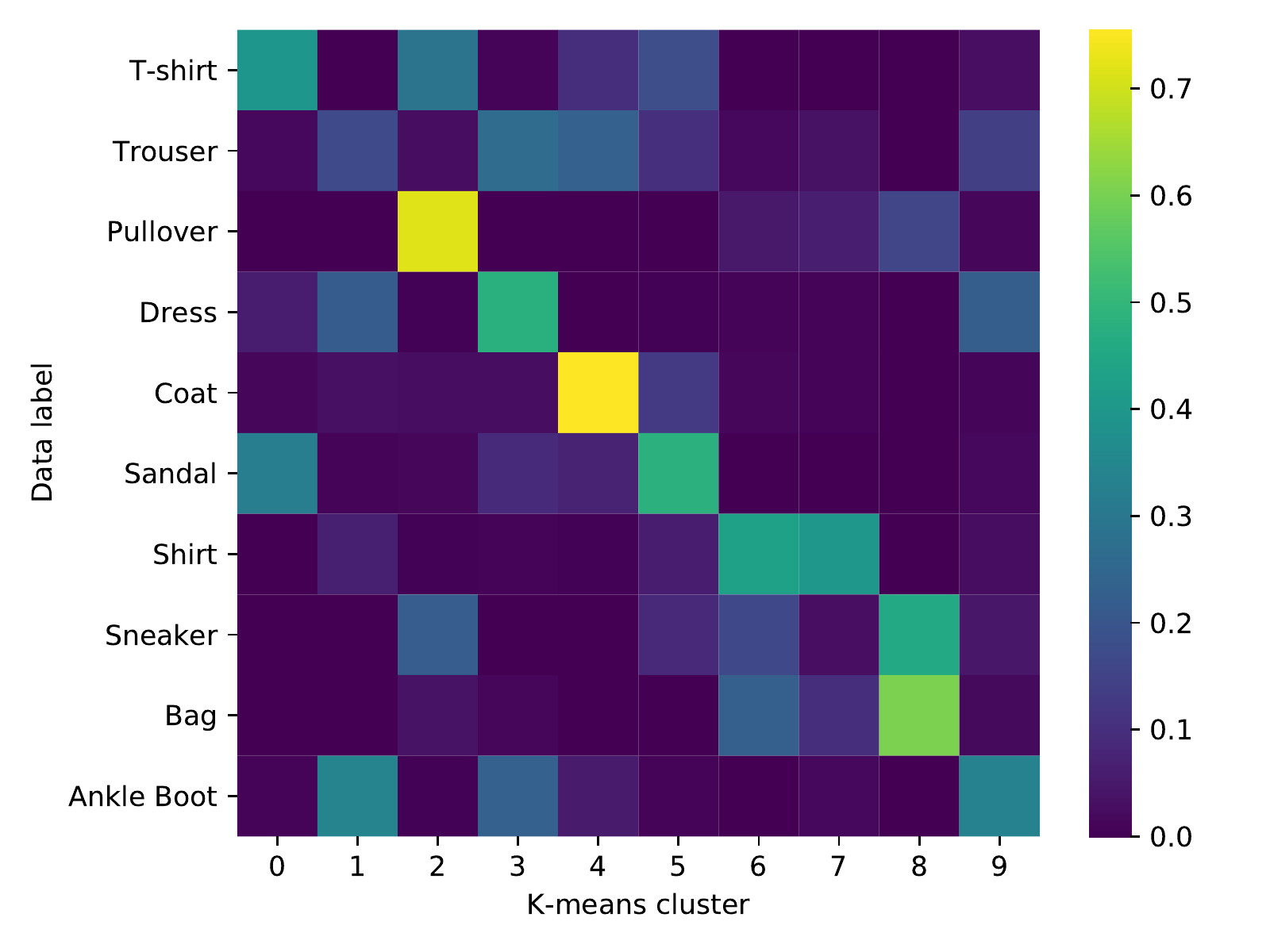}
         \caption{\centering}
         \label{fig:vae2dcluster}
     \end{subfigure}
\\~\\
     \begin{subfigure}[b]{0.33\linewidth}
         \includegraphics[width=\linewidth]{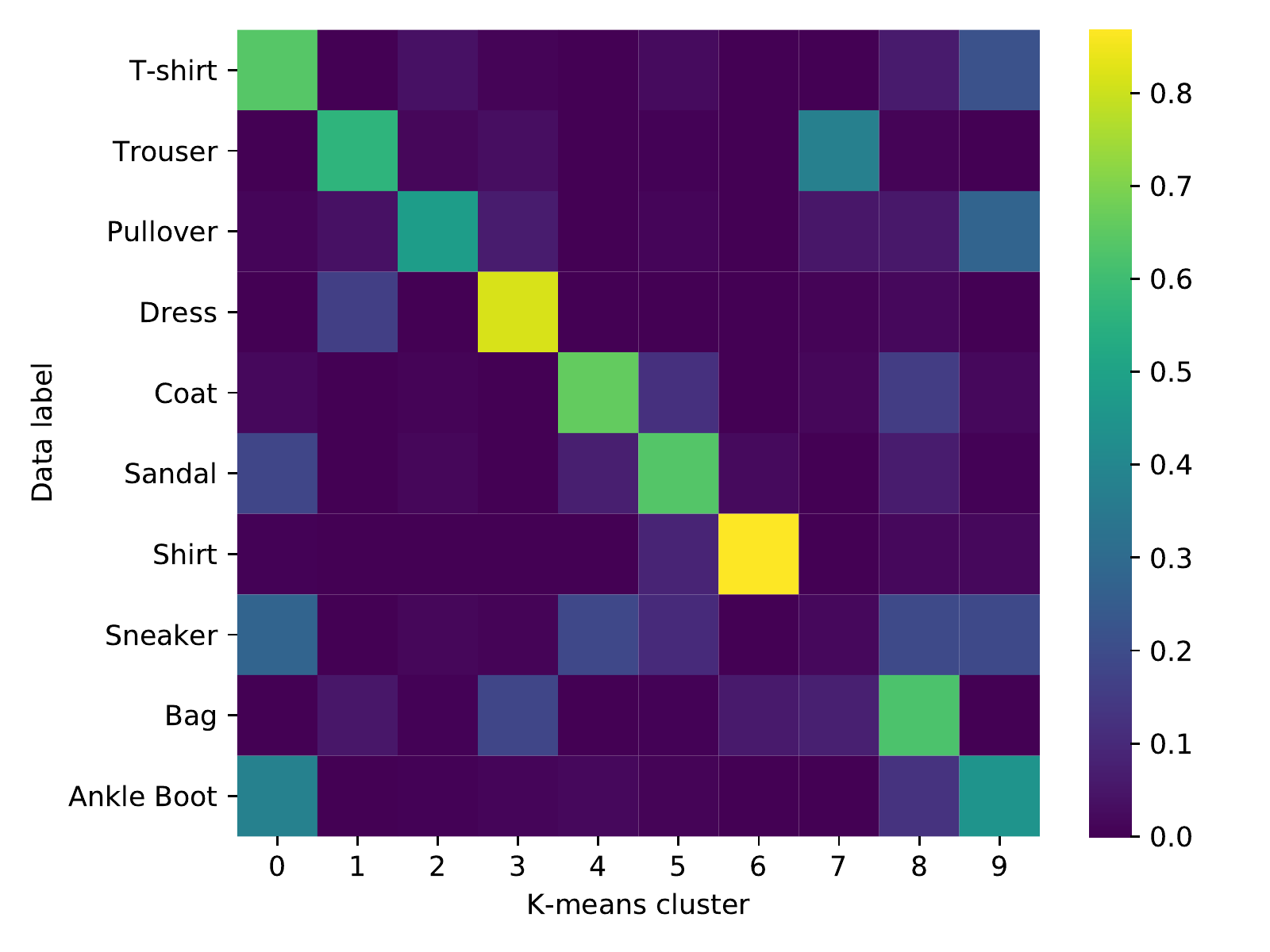}
         \caption{\centering}
         \label{fig:cvae2dcluster}
     \end{subfigure}
\caption{{(\textbf{a}) FisherNet. (\textbf{b}) VAE. (\textbf{c}) CVAE. Percentage overlap between the data labels and the clusters found by the K-means algorithm in a five-dimensional latent space.}}
\label{fig:5dcluster}
\end{figure}
\vspace{-12pt}
    \begin{specialtable}[H]\appendix
    \tablesize{\small}
\caption{ {Cluster trace.}}
\setlength{\cellWidtha}{\columnwidth/6-2\tabcolsep+0.0in}
\setlength{\cellWidthb}{\columnwidth/6-2\tabcolsep+0.0in}
\setlength{\cellWidthc}{\columnwidth/6-2\tabcolsep-0.0in}
\setlength{\cellWidthd}{\columnwidth/6-2\tabcolsep-0.0in}
\setlength{\cellWidthe}{\columnwidth/6-2\tabcolsep-0.0in}
\setlength{\cellWidthf}{\columnwidth/6-2\tabcolsep-0.0in}
\scalebox{1}[1]{\begin{tabularx}{\columnwidth}{>{\PreserveBackslash\raggedright}m{\cellWidtha}>{\PreserveBackslash\raggedright}m{\cellWidthb}>{\PreserveBackslash\raggedright}m{\cellWidthc}>{\PreserveBackslash\raggedright}m{\cellWidthd}>{\PreserveBackslash\raggedright}m{\cellWidthe}>{\PreserveBackslash\raggedright}m{\cellWidthf}}
\toprule
{\textbf{Latent Dimension} }& {\textbf{2} }& {\textbf{5} }& {\textbf{10} }& {\textbf{15} }& {\textbf{20}  }\\\midrule
{VAE              }& {4.849 }& {4.395 }& {4.088 }& {5.075 }& {5.622    }\\
{CVAE             }& {6.114 }& {5.752 }& {4.876 }& {4.516 }& {5.909    }\\
{FisherNet        }&  {5.504 }& {5.99  }& {6.08 }& {5.08 }& {5.319  }\\ 
\bottomrule
\end{tabularx}}
\label{tab:cluster}
\end{specialtable}
\vspace{-12pt}
\end{paracol}
\reftitle{References}


\externalbibliography{yes}

\end{document}